\documentclass[english,sigconf]{acmart}
\usepackage[T1]{fontenc}
\usepackage[latin9]{inputenc}
\usepackage{array}
\usepackage{float}
\usepackage{bbding}
\usepackage{url}
\usepackage{multirow}
\usepackage{amsmath}
\usepackage{graphicx}
\PassOptionsToPackage{normalem}{ulem}
\usepackage{ulem}

\makeatletter

\providecommand{\tabularnewline}{\\}

\copyrightyear{2023}
\acmYear{2023}
\setcopyright{acmlicensed}\acmConference[KDD '23]{Proceedings of the 29th ACM SIGKDD Conference on Knowledge Discovery and Data Mining}{August 6--10, 2023}{Long Beach, CA, USA}
\acmBooktitle{Proceedings of the 29th ACM SIGKDD Conference on Knowledge Discovery and Data Mining (KDD '23), August 6--10, 2023, Long Beach, CA, USA}
\acmPrice{15.00}
\acmDOI{10.1145/3580305.3599270}
\acmISBN{979-8-4007-0103-0/23/08}
\@ifundefined{showcaptionsetup}{}{%
 \PassOptionsToPackage{caption=false}{subfig}}
\usepackage{subfig}
\makeatother

\usepackage{babel}
\begin{document}
\title{Causal Inference via Style Transfer for Out-of-distribution Generalisation}
\author{Toan Nguyen}
\email{s222165627@deakin.edu.au} 
\orcid{0000-0003-2734-0622} 
\affiliation{\institution{Applied Artificial Intelligence Institute, Deakin University}\country{Australia}}

\author{Kien Do}
\email{k.do@deakin.edu.au} 
\orcid{0000-0002-0119-122X} 
\affiliation{\institution{Applied Artificial Intelligence Institute, Deakin University}   \country{Australia}}

\author{Duc Thanh Nguyen}
\email{duc.nguyen@deakin.edu.au} 
\orcid{0000-0002-2285-2066} 
\affiliation{\institution{School of Information Technology, Deakin University}   \country{Australia}}

\author{Bao Duong}
\email{duongng@deakin.edu.au} 
\orcid{0000-0001-9850-0270} 
\affiliation{\institution{Applied Artificial Intelligence Institute, Deakin University}   \country{Australia}}

\author{Thin Nguyen}
\email{thin.nguyen@deakin.edu.au} 
\orcid{0000-0003-3467-8963} 
\affiliation{\institution{Applied Artificial Intelligence Institute, Deakin University}   \country{Australia}}

\global\long\def\do{\mathrm{do}}%
\global\long\def\Model{\text{FAST}}%
\global\long\def\ModelFourier{\text{FAFT}}%
\global\long\def\ModelGeneral{\text{FAGT}}%
\global\long\def\argmax#1{\underset{#1}{\text{argmax }}}%
\global\long\def\argmin#1{\underset{#1}{\text{argmin }}}%

\begin{abstract}
Out-of-distribution (OOD) generalisation aims to build a model that
can generalise well on an unseen target domain using knowledge from
multiple source domains. To this end, the model should seek the causal
dependence between inputs and labels, which may be determined by the
semantics of inputs and remain invariant across domains. However,
statistical or non-causal methods often cannot capture this dependence
and perform poorly due to not considering spurious correlations learnt
from model training via unobserved confounders. A well-known existing
causal inference method like back-door adjustment cannot be applied
to remove spurious correlations as it requires the observation of
confounders. In this paper, we propose a novel method that effectively
deals with hidden confounders by successfully implementing front-door
adjustment (FA). FA requires the choice of a mediator, which we regard
as the semantic information of images that helps access the causal
mechanism without the need for observing confounders. Further, we
propose to estimate the combination of the mediator with other observed
images in the front-door formula via style transfer algorithms. Our
use of style transfer to estimate FA is novel and sensible for OOD
generalisation, which we justify by extensive experimental results
on widely used benchmark datasets.
\end{abstract}

\ccsdesc[500]{Computing methodologies~Causal reasoning and diagnostics} 
\ccsdesc[500]{Computing methodologies~Learning under covariate shift} 
\ccsdesc[300]{Computing methodologies~Computer vision}

\keywords{Causal Inference; Out-of-distribution Generalisation; Domain Generalisation; Style Transfer}

\maketitle

\section{Introduction\label{sec:Introduction}}

A common learning strategy in image classification is to maximise
the likelihood of the conditional distribution $P(Y|X)$ of a class
label $Y$ given an input image $X$ over a training set. This objective
is performed by exploiting the statistical dependence between input
images and class labels in the training set to yield good predictions
on a test set. Although this learning strategy works well for the
``in-distribution'' setting in which the test and training sets
are from the same domain, recent works~\cite{arjovsky2019invariant,lv2022causality,wang2022out}
have pointed out its limitations in dealing with out-of-distribution
(OOD) samples. This deficiency acts as a significant barrier to real-world
applications, where the test data may come from various domains and
may follow different distributions. 
\begin{figure}
\begin{centering}
\includegraphics[width=0.99\columnwidth]{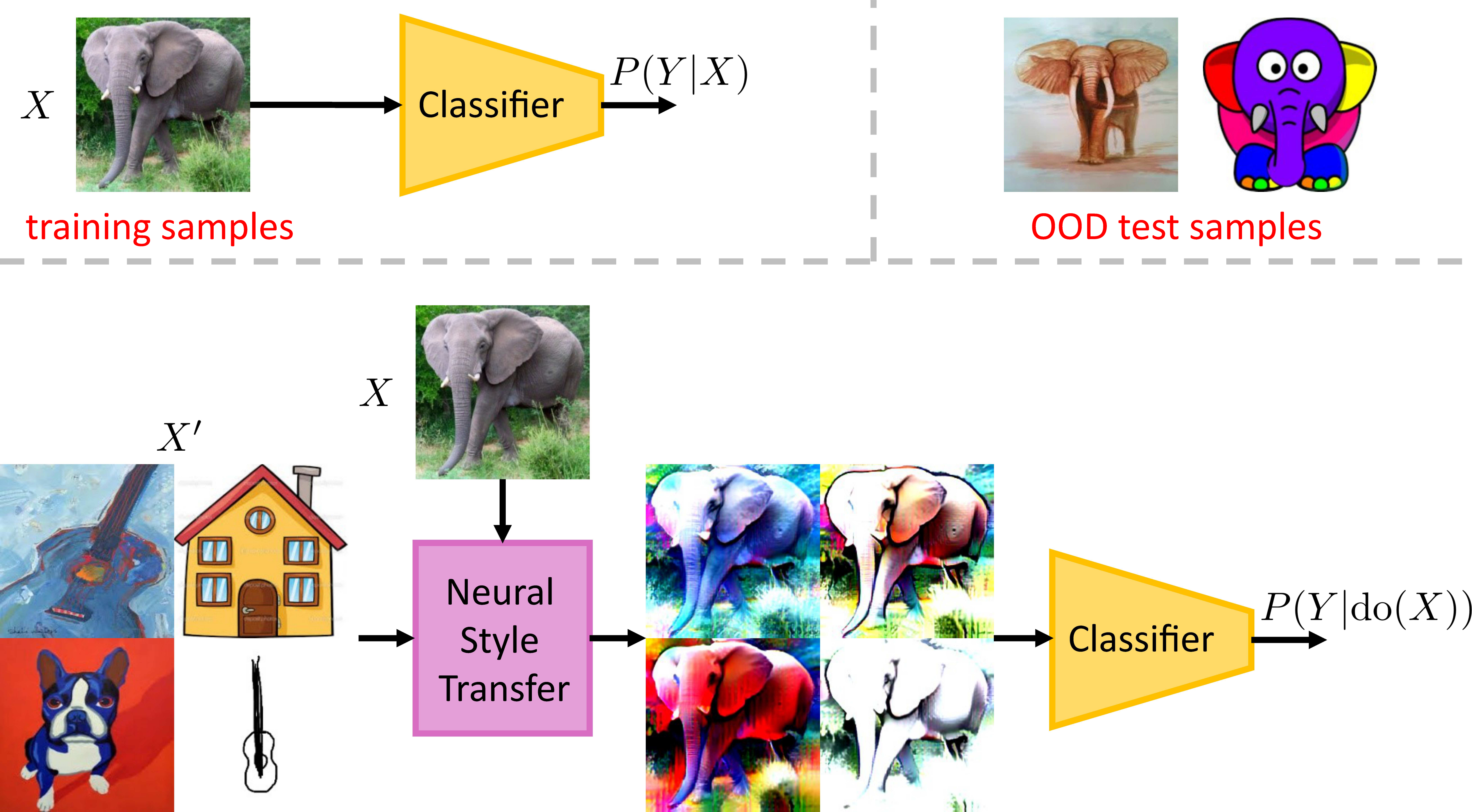} 
\par\end{centering}
\centering{}\caption{A model of $P(Y|X)$ (\textbf{top}) tends to capture all associations
between $X$ and $Y$ including the spurious dependence while a model
of $P(Y|\protect\do(X))$ (\textbf{bottom}) only captures the causal
dependence $X\rightarrow Y$. $P(Y|\protect\do(X))$ can be identified
via front-door adjustment by leveraging style transfer in our method.\label{fig:statistical_vs_causal}}
\end{figure}

The OOD generalisation task~\cite{shen2021towards,hendrycks2021many}
aims to address the issue by requiring deep learning models to infer
generalisable knowledge, which can be discovered by seeking the true
causation of labels or the causal dependence between inputs and labels~\cite{bernhard2021towards,scholkopf2022statistical,arjovsky2019invariant}.
However, owing to causal inference, statistical correlations in training
data do not always imply the true causation of features to labels~\cite{yao2021survey,glymour2016causal,scholkopf2022statistical}.
On the other hand, \emph{domain-specific nuisance features} that are
variant across domains can be learnt via statistical correlations~\cite{arjovsky2019invariant,deng2022deep,lv2022causality}.
These features are captured incidentally from confounding factors
such as background, or viewpoint, which change across domains and
confound the causal dependence between images and labels. For example,
when most training images of the elephant class are found with grass
(Fig.~\ref{fig:statistical_vs_causal}), by statistical learning,
a deep learning model is prone to taking into account features from
the grass (e.g., the green colour) as a signature for the elephant
class. This bias would mislead the model to wrong predictions on test
data from a different domain (e.g., when an elephant appears on a
stage or in a sketch painting). Here, by observing the statistical
dependence between the grass and the elephant class, one cannot conclude
either one of them is the cause of the other. Technically, a model,
which learns by the statistical correlation objective $P\left(Y|X\right)$
on a specific domain, might capture the spurious dependence induced
by confounders from that domain, leading to inferior generalisation
ability. 

Intuitively, we should model only the true causal dependence between
$X$ and $Y$. In causal inference literature, such dependence is
represented as $P(Y|\do(X))$. Here, the $\do(X)$ operator means
an intervention on $X$, which allows us to remove all spurious associations
between $X$ and $Y$~\cite{glymour2016causal,scholkopf2022statistical}.
$P(Y|\do(X))$ can be estimated via back-door adjustment~\cite{glymour2016causal}.
However, a well-known drawback of this approach is the need for observable
confounders. This requirement is not realistic and confounders are
also hard to be explicitly modelled (e.g., decomposing domain background
variables)~\cite{von2021self,glymour2016causal}. An alternative
approach to the issue is front-door adjustment (FA)~\cite{scholkopf2022statistical,glymour2016causal},
which supports causal inference with hidden confounders, making the
problem setting more practical. Implementation of FA requires the
choice of a mediator, which we regard as the \emph{semantic information
of images} that helps access the causal dependence, in other words,
features extracted from images that are \emph{always} predictive of
labels. Further, the front-door formula appeals to a feasible way
of modelling the combination of the class-predictive semantics of
an input image with the style and/or context of other images. Such
capability is naturally supported by style transfer models as illustrated
in Fig.~\ref{fig:statistical_vs_causal}. These methods have demonstrated
their ability to not only transfer artistic styles but also effectively
capture the semantic content of images~\cite{huang2017arbitrary,zhang2022exact,xu2021fourier},
which we propose that they hold promise for fulfilling the front-door
requirements. In this work, we propose a novel front-door adjustment-based
method to learn the causal distribution $P(Y|\do(X))$ in OOD generalisation.
We realise our method using two different style transfer models: AdaIN
Neural Style Transfer (NST)~\cite{huang2017arbitrary} and Fourier-based
Style Transfer (FST)~\cite{xu2021fourier}. Our use of style transfer
to estimate FA is novel and sensible for OOD generalisation, which
we show via the superior generalisation capacity of our model on widely
used benchmark datasets. In summary, our main contributions include: 
\begin{itemize}
\item A formulation of OOD generalisation under the perspective of causal
inference; 
\item A causal inference solution to OOD generalisation based on FA, which
we successfully implemented by two different style transfer strategies.
Our source code is available at:~\url{https://github.com/nktoan/Causal-Inference-via-Style-Transfer-for-OOD-Generalisation}.
\item Extensive experiments on various benchmark datasets, showing the superiority
of our proposed algorithm over state-of-the-art OOD generalisation
methods. 
\end{itemize}

\section{Related Work\label{sec:Related-Work}}

\textbf{Causal Inference }has been a fundamental research topic. It
aims to reconstruct causal relationships between variables and estimate
causal effects on variables of interest~\cite{yao2021survey,glymour2016causal}.
Causal inference has been applied to many deep learning research problems,
especially in domain adaptation (DA)~\cite{magliacane2018domain,rojas2018invariant,zhang2013domain,yue2021transporting}
and domain generalisation (DG)~\cite{wang2022out,mahajan2021domain,mitrovic2021representation}.
However, various difficulties curtail its development in deep learning.
The most common application of causal inference is removing spurious
correlations of domain-specific confounders that affect the generalisation
ability of deep models~\cite{yang2021causal,li2021confounder}. Existing
methods often adopt back-door adjustment to derive the causal dependence.
Examples include visual grounding~\cite{huang2022deconfounded},
semantic segmentation~\cite{zhang2020causal}, and object detection~\cite{wang2020visual}.
However, in practice, confounders may not be always observed. In addition,
they are hard to be derived~\cite{glymour2016causal,von2021self}.
In this case, front-door adjustment (FA)~\cite{pearl1995causal,glymour2016causal}
is often preferred. FA makes use of a mediator variable to capture
the intermediate causal effect. The effectiveness of the FA approach
has also been verified in many tasks such as vision-language tasks~\cite{yang2021causal},
image captioning~\cite{yang2021deconfounded}, and DG~\cite{li2021confounder}.
The main difference between our front-door-based algorithm and others
is the interpretation of the front-door formula in causal learning.
To the best of our knowledge, the incarnation of style transfer algorithms~\cite{huang2017arbitrary,jing2019neural,li2021random}
in FA is novel. It is a sensible and effective way for DG, which is
proven by our experimental results. Asides, to deal with hidden confounders,
instrumental~\cite{baiocchi2014instrumental,tang2021adversarial}
or proxy~\cite{yue2021transporting,miao2018identifying} variables
can also be used. However, in the DG problem, identifying the appropriate
mediator variable for FA presents fewer challenges and offers greater
practicality compared to selecting instrumental or proxy variables.
Causal inference methods are not straightforward to be technically
implemented for practical computer vision tasks. Our work can be seen
as the first attempt to effectively apply FA via style transfer algorithms
for DG.

\textbf{Domain Generalisation (DG)} aims to build image classification
models that can generalise well to unseen data. Typical DG methods
can be divided into three approaches including domain-invariant features
learning~\cite{muandet2013domain,peng2018synthetic}, meta-learning~\cite{zhao2021learning,li2020sequential,li2021confounder,dou2019domain},
and data/feature augmentation~\cite{zhou2021mixstyle,huang2020self,zhou2020learning}.
Domain-invariant features can be learnt from different source distributions
via adversarial learning~\cite{Li_2018_CVPR,li2018deep,motiian2017unified},
domain alignment~\cite{muandet2013domain,zhou2022domain}, or variational
Bayes~\cite{xiao2021bit}. Meta-learning addresses the DG problem
by simulating domain shift, e.g., by splitting data from source domains
into pseudo-training and pseudo-test data~\cite{li2018learning}.
Data/feature augmentation~\cite{zhou2020learning,wang2022out} can
be used to improve OOD generalisation by enhancing the diversity of
learnt features. Instead of seeking domain-invariant features, our
method aims to learn the causal mechanism between images and targets,
which is robust and invariant against domain shift~\cite{arjovsky2019invariant,li2021confounder,lv2022causality}.
Among data augmentation methods, our method is related to MixStyle~\cite{zhou2021mixstyle}
and EFDMix~\cite{zhang2022exact} due to the use of style transfer
(i.e., the AdaIN algorithm~\cite{huang2017arbitrary}) to create
novel data samples. However, our method considers preserving the content
of images as it is crucial in causal learning. Meanwhile, MixStyle
only interleaves some style transfer layers into image classification
models. In addition, both MixStyle and EFDMix rely on statistical
predictions, which are sensitive to domain shift~\cite{arjovsky2019invariant,mao2022causal}.
More, we regard the AdaIN algorithm as a tool to implement our proposed
front-door formula for estimating the causal dependence, resulting
in generalisable predictions. Similarly, the key advantage of our
method over FACT~\cite{xu2021fourier} is that we use the Fourier-based
transformation as an implementation of style transfer to infer the
causal dependence. 

\textbf{Causal Inference for DG}. The connection between causality
and DG has recently drawn considerable attention from the research
community. Much of the current line of work relies on the assumption
that there exists causal mechanisms or dependences between inputs
or input features and labels, which will be invariant across domains~\cite{teshima2020few,arjovsky2019invariant,mao2022causal,lv2022causality,duong2022bivariate}.
Many causality-inspired approaches~\cite{lv2022causality,wang2022out,mahajan2021domain,huang2023harnessing}
learn the domain-invariant representation or content feature that
can be seen as a direct cause of the label. Owing to the principle
of invariant or independent causal mechanisms\emph{~\cite{scholkopf2022statistical,peters2017elements,mitrovic2021representation},}
the causal mechanism between the content feature and the label remains
unchanged regardless of variations in other variables, such as domain
styles. Some methods~\cite{wang2022out,mitrovic2021representation,lv2022causality}
enforce regularisation with different data transformations on the
learnt feature to achieve the content feature. For example, CIRL~\cite{lv2022causality}
performs causal intervention via Fourier-based data augmentation~\cite{xu2021fourier}.
On the other hand, MatchDG~\cite{mahajan2021domain} assumes a causal
graph where the content feature is independent of the domain given
the hidden object and proposes a method that enforces this constraint
to the classifier by alternately performing contrastive learning and
finding good positive matches. One drawback of MatchDG is its reliance
on the availability of domain labels. Causal mechanism transfer~\cite{teshima2020few}
aims to capture causal mechanisms for DA via nonlinear independent
component analysis~\cite{hyvarinen2019nonlinear}. Asides, TCM~\cite{yue2021transporting}
leverages the proxy technique~\cite{yue2021transporting,miao2018identifying}
to perform causal inference for DA (where the target domain is provided)
rather than DG. 

In contrast, our method postulates that the causal dependence between
inputs and labels in DG can be identified by removing spurious correlations
induced by \emph{hidden} confounders~\cite{glymour2016causal}. On
the other hand, some methods~\cite{huang2023harnessing,lv2022causality}
assume that there is no confounding variable in their proposed causal
graph, which might be unlikely to be satisfied as proposed in many
previous methods~\cite{sun2021recovering,jiang2022invariant}. An
advantage of our approach over the above methods is that we do not
rely on imposing regularisation objectives, which may have gradient
conflicts with other learning objectives~\cite{chen2023pareto}.
Our method proposes to use the FA theorem~\cite{pearl1995causal,glymour2016causal}
and estimate the front-door formula via style transfer algorithms~\cite{huang2017arbitrary,jing2019neural,li2021random}.
Note that CICF~\cite{li2021confounder} also removes spurious correlations
using the FA theorem. However, while CICF deploys a meta-learning
algorithm to compute the front-door formula, we interpret the formula
via style transfer, which generates counterfactual images for causal
learning.

\section{Problem Formulation\label{sec:Problem_Formulation}}

Let $X$ and $Y$ be two random variables representing an input image
and its label, respectively. Statistical machine learning methods
for image classification typically model the observational distribution
$P\left(Y|X\right)$. They often perform well in the ``in-distribution''
(ID) setting where both training and test samples are from the same
domain. However, they may not generalise well in ``out-of-distribution''
(OOD) setting where test samples are from an unseen domain. As we
will show below, a possible reason for this phenomenon is that $P\left(Y|X\right)$
is \emph{domain-dependent} while a proper model for OOD generalisation
should be \emph{domain-invariant}.

\subsection{Out-of-Distribution Generalisation under the Causal Inference Perspective\label{subsec:Causal-Inference-Perspective}}

In order to clearly understand the limitation of statistical learning
in the OOD setting, we analyse the distribution $P\left(Y|X\right)$
under the causality perspective. In Fig.~\ref{fig:method_overview}(a),
we consider a causal graph $\mathcal{G}$ which describes possible
relationships between $X$ and $Y$ in the context of OOD generalisation.
In this graph, besides $X$ and $Y$, we also have two other random
variables $D$ and $U$ respectively specifying the \emph{domain-specific}
and \emph{domain-agnostic} features that influence both $X$ and $Y$.
$D$ and $U$ are unobserved since, in practice, their values are
rarely given during training and can only be inferred from the training
data. All connections between $X$ and $Y$ in $\mathcal{G}$ are
presented below. 
\begin{itemize}
\item $X\rightarrow Y:$ This is the \emph{true causal relationship} between
$X$ and $Y$ that describes how the \emph{semantics} in an image
$X$ lead to the prediction of a label $Y$. 
\item $X\leftarrow D\rightarrow Y:$ This is a \emph{spurious relationship}
showing that some \emph{domain-specific features} $D$ can act as
a confounder that distorts the true causal relationship between $X$
and $Y$. Different domains usually have different distributions for
$D$. As a result, if a model relies on $D$ to predict labels during
training, that model would generalise poorly to an OOD test domain.
An example of this confounding dependence is that most ``giraffe''
images from the ``art-painting'' domain are drawn with a specific
texture, and a model trained on these samples may associate the texture
with the class ``giraffe''. 
\item $X\leftarrow U\rightarrow Y:$ This is another \emph{spurious} \emph{relationship}
specifying the confounding dependence between $X$ and $Y$ caused
by \emph{domain-agnostic features} $U$. For example, since most elephants
are gray, a model may misuse this colour information to predict a
dog in gray as an elephant. Here, ``gray'' is an inherent property
of elephants and may not depend on domains. 
\end{itemize}
According to the graph $\mathcal{G}$, the statistical distribution
$P\left(Y|X=x\right)$ has the following mathematical expression:
\begin{align}
P\left(Y|x\right) & =\sum_{d,u}p\left(d,u|x\right)P\left(Y|x,d,u\right)\\
 & =\mathbb{E}_{p\left(d,u\mid x\right)}\left[P\left(Y|x,d,u\right)\right]
\end{align}
where $d$, $u$ denote specific values of $D$, $U$ respectively.
Here, we make a slight notation abuse by using $\sum$ as an integral
operator. We can see that $P\left(Y|x\right)$ is comprised of the
\emph{causal and spurious relationships }which correspond to $x$,
$d$ and $u$ in $P\left(Y|x,d,u\right)$. Since $p(d,u|x)$ can be
different for different $x$, a model of $P\left(Y|x\right)$ is likely
biased towards some domain-specific and domain-agnostic features $d$,
$u$. Therefore, in an unseen test domain, $P(Y|x)$ tends to make
wrong label predictions.

Naturally, we are interested in a model that \emph{captures only the
true causal relationship} between $X$ and $Y$. From the perspective
of causal inference, we should model $P\left(Y|\do(x)\right)$ rather
than $P\left(Y|x\right)$. $\do(x)$ or $\do(X=x)$ is a notation
from the do-calculus~\cite{glymour2016causal} meaning that we intervene
$X$ by setting its value to $x$ regardless of the values of $U$
and $D$. Graphically, the intervention $\do(x)$ removes all incoming
links from $D$, $U$ to $X$, blocking all associative paths from
$X$ to $Y$ except for the direct causal path $X\rightarrow Y$.
$P\left(Y|\do(x)\right)$ is, therefore, regarded as the distribution
of the potential outcome $Y$ under the intervention $\do(x)$.

\paragraph{Maximum Interventional Log-likelihood\label{par:Maximum-Interventional-LogLikelihood}}

If we can devise a parametric model $P_{\theta}\left(Y|\do(x)\right)$
of $P\left(Y|\do(x)\right)$ using observational training data $\mathcal{D}$,
we can train this model by optimizing the following objective:

\begin{align}
\theta^{*} & =\argmax{\theta}\mathbb{E}_{(x,y)\sim\mathcal{D}}\left[\log P_{\theta}\left(Y=y|\do(x)\right)\right]\label{eq:MCL}
\end{align}
We refer to the objective in Eq.~(\ref{eq:MCL}) as \emph{Maximum
Interventional Log-likelihood (MIL)} to distinguish it from the conventional
Maximum Log-likelihood (ML) for $P\left(Y|x\right)$. We note that
Eq.~(\ref{eq:MCL}) makes use of the \emph{consistency} and \emph{no-interference}
\emph{assumptions~\cite{yao2021survey}} in causal inference. The
former states that the observed outcome $y$ of an observed treatment
$X=x$ is the potential outcome of $Y$ under the intervention $\do(X=x)$,
and the later states that the potential outcome $Y_{i}$ of an item
(or sample) $i$ is only affected by the intervention on this item
(or $\do(X_{i}=x)$). Detailed derivations of Eq.~(\ref{eq:MCL})
are provided in the supplementary material. For the ease of notation,
we use $P\left(Y|\do(x)\right)$ hereafter to denote a parametric
model of the potential outcome $Y$.

\subsection{Identification via Back-door Adjustment\label{subsec:Backdoor-Adj}}

Given the appeal of modelling $P\left(Y|\do(x)\right)$ for OOD generalisation,
the next question is ``Can we identify this causal distribution?'',
or more plainly, ``Can we express $P\left(Y|\do(x)\right)$ in terms
of conventional statistical distributions?''. A common approach to
identify $P\left(Y|\do(x)\right)$ is using the \emph{back-door adjustment}
as below: 
\begin{equation}
P\left(Y|\do(x)\right)=\sum_{d,u}p\left(d,u\right)P\left(Y|x,d,u\right)\label{eq:backdoor_adj}
\end{equation}
However, this back-door adjustment is only applicable if $D$ and
$U$ are observed. In the case $D$ and $U$ are unobserved as specified
in the graph $\mathcal{G}$, there is no way to estimate $p(d,u)$
in Eq.~(\ref{eq:backdoor_adj}) without using $x$. Therefore, there
is a need for another approach to identify $P\left(Y|\do(x)\right)$.

\section{Proposed Method\label{sec:Method}}

We present our proposed front-door adjustment-based method for modelling
$P\left(Y|\do(x)\right)$ to deal with hidden confounders in Section~\ref{subsec:Frontdoor-Adjustment}
and introduce two different ways to estimate it for OOD image classification
in Section~\ref{subsec:Neural-Style-Transfer} and Section~\ref{subsec:Generalisation-to-other}.

\begin{figure*}
\noindent %
\noindent\begin{minipage}[b]{0.21\textwidth}%
\begin{tabular}[b]{c}
\includegraphics[width=1\textwidth]{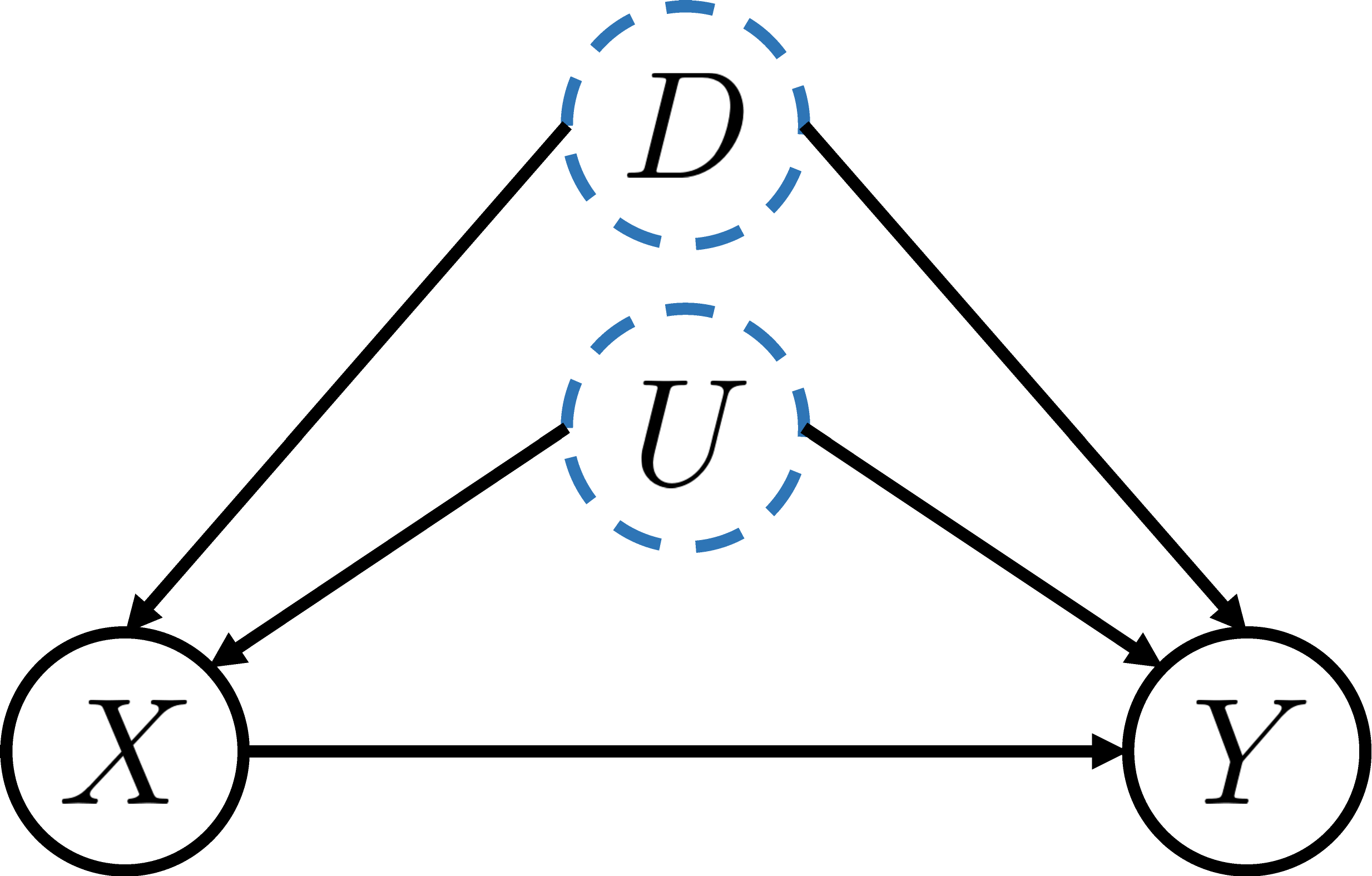}\tabularnewline
\label{(a)causal-graph}(a)\tabularnewline
\tabularnewline
\includegraphics[width=1\textwidth]{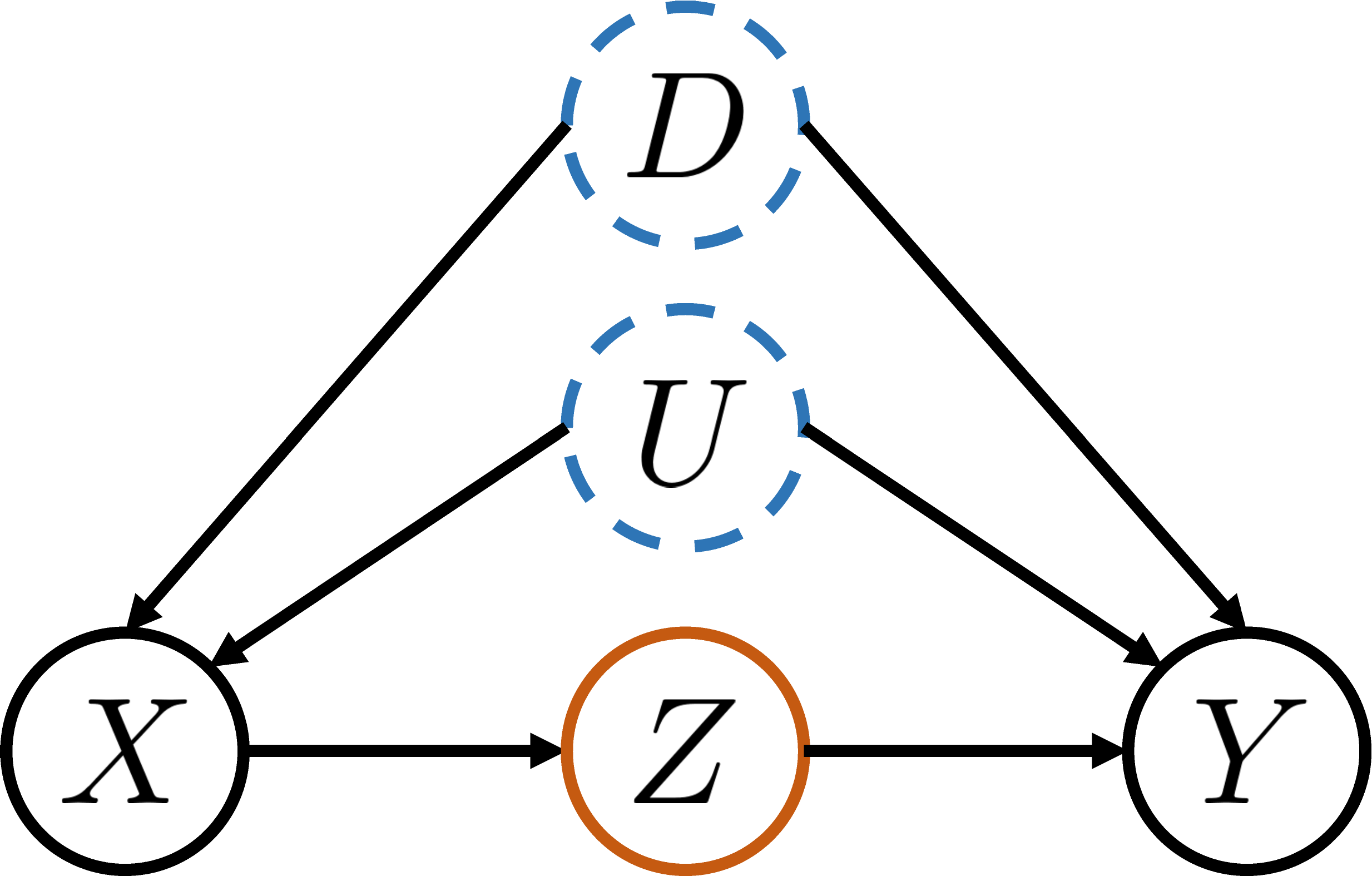}\tabularnewline
(b)\tabularnewline
\end{tabular}%
\end{minipage}\hspace{0.06\textwidth}%
\noindent\begin{minipage}[b]{0.71\textwidth}%
\begin{tabular}[b]{c}
\includegraphics[width=1\textwidth]{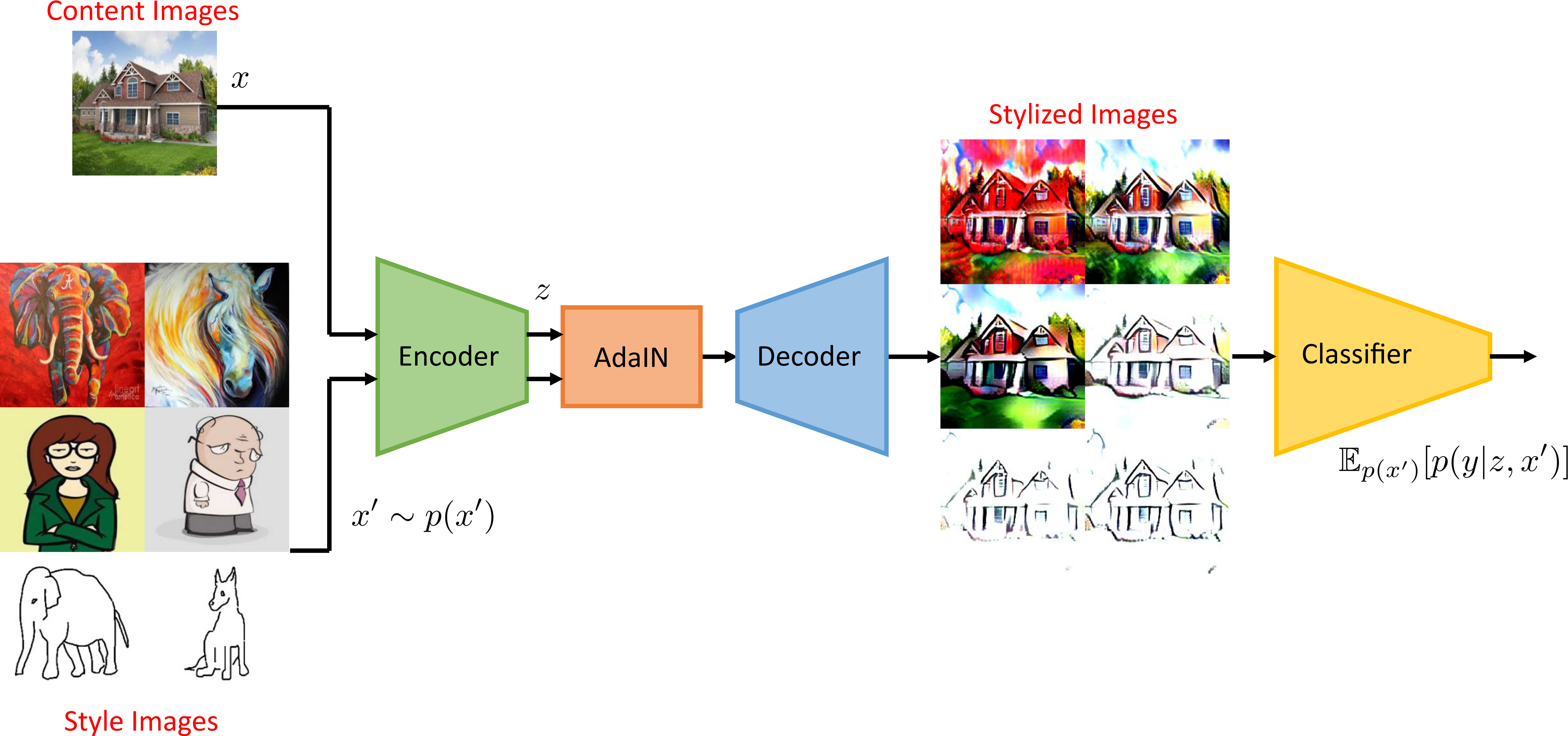}\tabularnewline
\label{fig:overview}(c)\tabularnewline
\end{tabular}%
\end{minipage}

\noindent \caption{\textbf{(a)}: A causal graph $\mathcal{G}$ describing the \emph{causal
relationship}, \emph{domain-specific association} (via $D$), and
\emph{domain-agnostic association} (via $U$) between an input image
$X$ and a label $Y$. Variables with dashed border ($D$, $U$) are
unobserved. \textbf{(b)}: A variant of $\mathcal{G}$ with a mediator
$Z$ between $X$ and $Y$ that enables identification of $P\left(Y|\protect\do(x)\right)$
via front-door adjustment. \textbf{(c)}: Illustration of our proposed
method that leverages style transfer models~\cite{huang2017arbitrary,xu2021fourier}
to perform front-door adjustment.\label{fig:method_overview}}
\end{figure*}

\subsection{Identification via Front-door Adjustment\label{subsec:Frontdoor-Adjustment}}

In order to use front-door adjustment, we assume that there is a random
variable $Z$ acting as a mediator between $X$ and $Y$ as illustrated
in Fig.~\ref{fig:method_overview}(b). $Z$ must satisfy three conditions
as follows~\cite{glymour2016causal}: (i) All directed paths from
$X$ to $Y$ flow through $Z$.~(ii) $X$ blocks all back-door paths
from $Z$ to $Y$.~(iii) There are no unblocked back-door paths from
$X$ to $Z$.

The first two conditions suggest that $Z$ should capture all and
only semantic information in $X$ that is always predictive of $Y$,
while the last condition means that we should be able to estimate
the causal relationship $X\rightarrow Z$ by using just the observed
training images. Such variable $Z$ can be captured by several learning
methods, such as style transfer models~\cite{huang2017arbitrary,jing2019neural}
or contrastive learning~\cite{von2021self,mahajan2021domain}. Even
if the assumptions do not hold perfectly in practice and the mediator
captures some spurious features in $X$, our method simply reverts
to the standard empirical risk minimisation (ERM) model~\cite{Vapnik1998},
which is widely applicable and can be used on any dataset or problem.
Importantly, the more semantically meaningful information that the
mediator $Z$ captures from $X$, the better our method can approximate
$P\left(Y|\do(x)\right)$. We refer the reader to our supplementary
material for a proof sketch that $Z$ as semantic features will satisfy
all the front-door criterions. Given the above conditions, we can
determine $P\left(Y|\do(x)\right)$ via front-door adjustment as follows:
\begin{align}
P\left(Y|\do(x)\right) & =\sum_{z}p(z|\do(x))P(Y|z,\do(x))\\
 & =\sum_{z}p(z|x)P(Y|\do(z),\do(x))\\
 & =\sum_{z}p(z|x)P(Y|\do(z))\\
 & =\sum_{z}p(z|x)\left(\sum_{x'}p(x')P(Y|z,x')\right)\\
 & =\mathbb{E}_{p(z|x)}\mathbb{E}_{p(x')}[P(Y|z,x')]\label{eq:frontdoor_adj_final}
\end{align}
where $z$, $x'$ denote specific values of $Z$ and $X$, respectively.
Detailed derivations of all steps in Eq.~(\ref{eq:frontdoor_adj_final})
are provided in the supplementary material.

One interesting property of the final expression in Eq.~(\ref{eq:frontdoor_adj_final})
is that the label prediction probability $P(Y|z,x')$ no longer depends
on the domain-specific and domain-agnostic features $D$ and $U$
unlike $P(Y|x,d,u)$ in the back-door adjustment formula in Eq.~(\ref{eq:backdoor_adj}).
It suggests that by using front-door adjustment, we can explicitly
handle the OOD generalisation problem. In addition, the final expression
is elegant as the two expectations can be easily estimated via Monte
Carlo sampling and $p(z|x)$ can be easily modelled via an encoder
that maps $x$ to $z$. The only remaining problem is modelling $P(Y|z,x')$.
In the next section, we propose a novel way to model $P(Y|z,x')$
in front-door adjustment by leveraging neural style transfer.

\subsection{Front-door Adjustment via Neural Style Transfer\label{subsec:Neural-Style-Transfer}}

The main difficulty of modelling $P(Y|z,x')$ is designing a network
that can combine the semantic features $z$ of an image $x$ with
another image $x'$ without changing the class-predictive semantic
information in $z$. Here, we want a model that can fuse the content
of $x$ with the style of $x'$. A family of models that has such
capability is neural style transfer (NST). Within the scope of this
paper, we consider AdaIN NST~\cite{huang2017arbitrary} as a representative
NST method and use it to model $P(Y|z,x')$. We note that it is straightforward
to use other NST methods, e.g.,~\cite{zhang2022exact} in place of
AdaIN NST. We refer this version to as $\Model$ (\textbf{\emph{F}}\emph{ront-door
}\textbf{\emph{A}}\emph{djustment via }\textbf{\emph{S}}\emph{tyle
}\textbf{\emph{T}}\emph{ransfer}). Below, we describe AdaIN NST and
the training objective of our method.

An AdaIN NST model consists of an encoder $E$, an AdaIN module, and
a decoder $D$ as shown in Fig.~\ref{fig:method_overview}(c). The
encoder $E$ maps a content image $x$ and a style image $x'$ to
their corresponding feature vectors $z=E(x)$ and $z'=E(x')$, respectively.
The content and style features $z$, $z'$ then serve as inputs to
the AdaIN module to produce a \emph{stylised feature} $\tilde{z}$
as follows: 
\begin{equation}
\tilde{z}=\text{AdaIN}(z,z')=\mu(z')+\sigma(z')\frac{z-\mu(z)}{\sigma(z)}
\end{equation}
where $\mu(z)$, $\sigma(z)$ denote the mean and standard deviation
vectors of $z$ along its channel; so as $\mu(z')$, $\sigma(z')$
for $z'$. Specifically, if $z$ is a feature map of shape $C\times H\times W$
with $C$, $H$, $W$ being the number of channels, height, and width,
then $\mu(z)$, $\sigma(z)$ are computed as follows: 
\begin{align}
\mu_{c}(z) & =\frac{1}{HW}\sum_{h=1}^{H}\sum_{w=1}^{W}z_{c,h,w},\\
\sigma_{c}(z) & =\sqrt{\frac{1}{HW-1}\sum_{h=1}^{H}\sum_{w=1}^{W}\left(z_{c,h,w}-\mu_{c}(z)\right)^{2}}
\end{align}
For better controlling the amount of style transferred from $x'$
to $x$, we perform a linear interpolation between $\tilde{z}$ and
$z$, then set the result back to $\tilde{z}$ as follows: 
\begin{equation}
\tilde{z}\leftarrow\alpha\tilde{z}+(1-\alpha)z\label{eq:stylized_z_itpl}
\end{equation}
where $0\leq\alpha\leq1$ is the interpolation coefficient. The stylised
code $\tilde{z}$ in Eq.~(\ref{eq:stylized_z_itpl}) is sent to the
decoder $D$ to generate a \emph{stylised image} $\tilde{x}=D(\tilde{z})$.
In summary, we can write $\tilde{x}$ as a function of $x$ and $x'$
in a compact form below:

\begin{equation}
\tilde{x}=\text{AdaIN\_NST}(x,x')\label{eq:AdaIN_NST}
\end{equation}
where AdaIN\_NST$(\cdot,\cdot)$ denotes the AdaIN NST model described
above. We can easily substitute AdaIN\_NST by other NST methods to
compute $\tilde{x}$ from $x$ and $x'$.

Once the AdaIN NST model has been learnt well, $\tilde{x}$ will capture
the semantic feature $z$ of $x$ and the style of $x'$ properly.
Thus, we can condition $Y$ on $\tilde{x}$ instead of $z$ and $x'$
and regard $\mathbb{E}_{p(x')}\left[P(Y|\tilde{x})\right]$ as an
approximation of $\mathbb{E}_{p(z|x)}\mathbb{E}_{p(x')}\left[P(Y|z,x')\right]$.
We compute the label of $\tilde{x}$ by feeding this stylised image
through the classifier $F$.

We train $F$ using the Maximum Interventional Likelihood criterion
described in Section~\ref{subsec:Causal-Inference-Perspective}.
The loss w.r.t. each training sample $(x,y)$ is given below: 
\begin{align}
\mathcal{L}_{\text{\ensuremath{\Model}}}(x,y) & =-\log P\left(Y=y|\do(x)\right)\\
 & =\mathcal{L}_{\text{xent}}\left(\mathbb{E}_{p(x')}\left[\left(F(\tilde{x})\right)\right],y\right)\label{eq:Loss_causal_2}\\
 & =\mathcal{L}_{\text{xent}}\left(\left(\frac{1}{K}\sum_{k=1}^{K}F\left(\tilde{x}_{k}\right)\right),y\right)\label{eq:Loss_causal_3}
\end{align}
where $F(\cdot)$ denotes the class probability produced by $F$ for
an input image, $\mathcal{L}_{\text{xent}}$ denotes the cross-entropy
loss, and $\tilde{x}$ is derived from $x$ and $x'$ via Eq.~(\ref{eq:AdaIN_NST}).
The RHS of Eq.~(\ref{eq:Loss_causal_3}) is the Monte Carlo estimate
of the counterpart in Eq.~(\ref{eq:Loss_causal_2}) with $K$ the
number of style images $x'_{1},...,x'_{K}$ and $\tilde{x}_{k}=\text{AdaIN\_NST}(x,x'_{k})$.
Due to the arbitrariness of the style images $\{x'\}$, the stylised
images $\{\tilde{x}\}$ can have large variance despite sharing the
same content $x$. We found that the large variance of $\{\tilde{x}\}$
can make the learning fluctuating. To avoid this, we combine the label
predictions of $K$ stylised images with that of the content image,
and use a loss: 
\begin{align}
\mathcal{L_{\text{\ensuremath{\Model}}}}(x,y) & =\mathcal{L}_{\text{xent}}\left(\left(\beta F(x)+\frac{1-\beta}{K}\sum_{k=1}^{K}F\left(\tilde{x}_{k}\right)\right),y\right)\label{eq:final_loss-FAST}
\end{align}
where $0\leq\beta<1$ is the interpolation coefficient between $F(x)$
and $\frac{1}{K}\sum_{k=1}^{K}F(\tilde{x}_{k})$.

\subsection{Front-door Adjustment via Fourier-based Style Transfer\label{subsec:Generalisation-to-other}}

Our proposed method for front-door adjustment can seamlessly generalise
to other style transfer methods. In this paper, we also consider the
\emph{Fourier-based Style Transfer} (FST) used in~\cite{lv2022causality,xu2021fourier}.
This style transfer method applies the discrete Fourier transform
to decompose an image into its phase and amplitude, that are considered
as content and style, respectively. Unlike NST, FST does not require
a training phase. However, its decomposed styles may not as diverse
as those learnt from NST.

The Fourier transformation $\mathcal{F}$ of an image $x$ can be
written $\mathcal{F}(x)=\mathcal{A}(x)\times e^{-j\times\mathcal{P}(x)}$,
where $\mathcal{A}(x)$ and $\mathcal{P}(x)$ denote the amplitude
and phase of $x$, respectively. Following~\cite{xu2021fourier},
we use the \emph{``amplitude mix''} strategy to produce a stylised
image $\hat{x}$ from a content image $x$ and a style image $x'$
as follows: 
\begin{align}
\hat{x} & =\mathcal{F}^{-1}\left(\left((1-\lambda)\mathcal{A}(x)+\lambda\mathcal{A}(x')\right)\times e^{-j\times\mathcal{P}(x)}\right)
\end{align}
where $\lambda\sim U(0,\eta)$ and $0\leq\eta\leq1$ is a hyper-parameter
controlling the maximum style mixing rate.

We name this version as $\ModelFourier$ (\textbf{F}ront-door \textbf{A}djustment
via \textbf{F}ourier-based\textbf{ }Style \textbf{T}ransfer)\emph{.
}We train the classifier $F$ using the same loss function as in Eq.~(\ref{eq:final_loss-FAST})
but with $\tilde{x}_{k}$ being replaced by $\hat{x}_{k}$ which is
given below:

\begin{align}
\mathcal{L}_{\ModelFourier}(x,y) & =\mathcal{L}_{\text{xent}}\left(\left(\beta F(x)+\frac{1-\beta}{K}\sum_{k=1}^{K}F\left(\hat{x}_{k}\right)\right),y\right)\label{eq:final_loss-FAFT}
\end{align}

Since different style transfer approaches capture different styles
in the input, one can combine NST and FST together to exploit the
strengths of both methods. This results in a more general manner namely
$\ModelGeneral$ (\textbf{\emph{F}}\emph{ront-door }\textbf{\emph{A}}\emph{djustment
via }\textbf{\emph{G}}\emph{eneral }\textbf{\emph{T}}\emph{ransfer}),
where the classification network $F$ is optimised by the following
loss function:

\begin{align}
\mathcal{L}_{\ModelGeneral}(x,y) & =\mathcal{L}_{\text{xent}}\left(\left(\beta F(x)+\frac{1-\beta}{2K}\sum_{k=1}^{K}\left[F\left(\tilde{x}_{k}\right)+F\left(\hat{x}_{k}\right)\right]\right),y\right)\label{eq:final_loss-FAGT}
\end{align}

\subsection{Inference by the Causal Dependence}

We use the causal dependence $P(Y|do(X))$ not only for training but
also for the inference phase. The association dependence $P(Y|X)$
could bring spuriousness into predictions, even if the model is trained
with the causal objective. In both phases, for all methods, we sample
$K=3$ training samples from different domains\emph{ }to compute $P(Y|\do(X))$.
We empirically show the effectiveness of using $P(Y|do(X))$ over
$P(Y|X)$ in both the two phases in Section~\ref{sec:Experiments},
where the method with causality in both phases clearly outperforms
the others. It is worth noting that for every single image $x$, we
do not use its neighbors in the representation space as a substitution
for either $\tilde{x}$ or $\hat{x}$, since the specific semantics
of $x$ in these samples would not suffice to infer $P(Y|do(X=x))$.

\section{Experiments\label{sec:Experiments}}

\subsection{Datasets}

\begin{table*}
\begin{centering}
\resizebox{.99\textwidth}{!}{{\small{}{}{}}%
\begin{tabular}{c|cccc|c||cccc|c||cccc|c}
\hline 
\multirow{2}{*}{Methods} & \multicolumn{5}{c||}{PACS} & \multicolumn{5}{c||}{Office-Home} & \multicolumn{5}{c}{Digits-DG}\tabularnewline
\cline{2-16} \cline{3-16} \cline{4-16} \cline{5-16} \cline{6-16} \cline{7-16} \cline{8-16} \cline{9-16} \cline{10-16} \cline{11-16} \cline{12-16} \cline{13-16} \cline{14-16} \cline{15-16} \cline{16-16} 
 & A & C & P & S & \emph{Avg.} & A & C & P & R & \emph{Avg.} & MN & MM & SV & SY & \emph{Avg.}\tabularnewline
\hline 
\hline 
ERM~\cite{Vapnik1998} & 77.0 & 75.9 & 96.0 & 69.2 & 79.5 & 58.1 & 48.7 & 74.0 & 75.6 & 64.2 & 95.8 & 58.8 & 61.7 & 78.6 & 73.7\tabularnewline
\hline 
\hline 
DeepAll~\cite{zhou2020deep} & 77.6 & 76.8 & 95.9 & 69.5 & 79.9 & 57.9 & 52.7 & 73.5 & 74.8 & 64.7 & 95.8 & 58.8 & 61.7 & 78.6 & 73.7\tabularnewline
Jigen~\cite{Carlucci_2019_CVPR} & 79.4 & 75.3 & 96.0 & 71.4 & 80.5 & 53.0 & 47.5 & 71.5 & 72.8 & 61.2 & 96.5 & 61.4 & 63.7 & 74.0 & 73.9\tabularnewline
CCSA~\cite{motiian2017unified} & 80.5 & 76.9 & 93.6 & 66.8 & 79.4 & 59.9 & 49.9 & 74.1 & 75.7 & 64.9 & 95.2 & 58.2 & 65.5 & 79.1 & 74.5\tabularnewline
MMD-AAE~\cite{Li_2018_CVPR} & 75.2 & 72.7 & 96.0 & 64.2 & 77.0 & 56.5 & 47.3 & 72.1 & 74.8 & 62.7 & 96.5 & 58.4 & 65.0 & 78.4 & 74.6\tabularnewline
CrossGrad~\cite{shankar2018generalizing} & 79.8 & 76.8 & 96.0 & 70.2 & 80.7 & 58.4 & 49.4 & 73.9 & 75.8 & 64.4 & 96.7 & 61.1 & 65.3 & 80.2 & 75.8\tabularnewline
DDAIG~\cite{zhou2020deep} & 84.2 & 78.1 & 95.3 & 74.7 & 83.1 & 59.2 & 52.3 & \uline{74.6} & \uline{76.0} & 65.5 & 96.6 & 64.1 & 68.6 & 81.0 & 77.6\tabularnewline
L2A-OT~\cite{zhou2020learning} & 83.3 & 78.2 & 96.2 & 73.6 & 82.8 & \textbf{60.6} & 50.1 & \textbf{74.8} & \textbf{77.0} & 65.6 & 96.7 & 63.9 & 68.6 & 83.2 & 78.1\tabularnewline
RSC~\cite{huang2020self} & 83.4 & 80.3 & 96.0 & 80.9 & 85.2 & 58.4 & 47.9 & 71.6 & 74.5 & 63.1 & - & - & - & - & -\tabularnewline
MixStyle~\cite{zhou2021mixstyle} & 83.0 & 78.6 & 96.3 & 71.2 & 82.3 & 58.7 & 53.4 & 74.2 & 75.9 & 65.5 & 96.5 & 63.5 & 64.7 & 81.2 & 76.5\tabularnewline
FACT~\cite{xu2021fourier} & 85.4 & 78.4 & 95.2 & 79.2 & 84.6 & 60.3 & 54.9 & 74.5 & 76.6 & 66.6 & 97.9 & 65.6 & 72.4 & 90.3 & 81.6\tabularnewline
FACT$^{\dagger}$ & 84.8 & 78.0 & 95.3 & 79.1 & 84.3 & 57.9 & 54.9 & 73.7 & 75.0 & 65.4 & 95.6 & 63.3 & 69.0 & 82.5 & 77.6\tabularnewline
\hline 
\hline 
MatchDG~\cite{mahajan2021domain} & 81.3 & \uline{80.7} & 96.5 & 79.7 & 84.6 & - & - & - & - & - & - & - & - & - & -\tabularnewline
CICF~\cite{li2021confounder} & 80.7 & 76.9 & 95.6 & 74.5 & 81.9 & 57.1 & 52.0 & 74.1 & 75.6 & 64.7 & 95.8 & 63.7 & 65.8 & 80.7 & 76.5\tabularnewline
CIRL~\cite{lv2022causality} & 86.1 & 80.6 & 95.9 & 82.7 & 86.3 & 61.5 & 55.3 & 75.1 & 76.6 & 67.1 & 96.1 & 69.8 & 76.2 & 87.7 & 82.5\tabularnewline
CIRL$^{\dagger}$ & 85.9 & 79.8 & 95.6 & \uline{81.4} & 85.7 & 58.6 & \textbf{55.4} & 73.8 & 75.1 & 65.7 & 95.4 & \textbf{65.9} & 68.9 & 82.0 & 78.1\tabularnewline
\hline 
\hline 
$\Model$ (ours) & \uline{86.5} & \uline{80.7} & \uline{96.6} & 81.0 & \uline{86.2} & 59.6 & 54.5 & 74.1 & 75.7 & \uline{66.0} & \uline{97.9} & 64.0 & \uline{70.5} & \uline{90.0} & \uline{80.6}\tabularnewline
$\ModelFourier$ (ours) & 86.3 & 80.3 & 96.4 & 80.8 & 86.0 & 60.0 & 53.6 & 73.9 & 75.7 & 65.8 & 97.8 & 62.9 & 68.5 & 89.6 & 79.7\tabularnewline
$\ModelGeneral$ (ours) & \textbf{87.5} & \textbf{80.9} & \textbf{96.9} & \textbf{81.9} & \textbf{86.8} & \uline{60.1} & \uline{55.0} & 74.5 & 75.8 & \textbf{66.4} & \textbf{98.3} & \uline{65.7} & \textbf{70.9} & \textbf{90.4} & \textbf{81.3}\tabularnewline
\hline 
\end{tabular}}
\par\end{centering}
\centering{}\caption{Leave-one-domain-out classification accuracies (in \%) on PACS, Office-Home,
and Digits-DG. In ERM, we simply train the classifier on all domains
except the test one. FACT$^{\dagger}$ and CIRL$^{\dagger}$ denote
our rerun of FACT and CIRL respectively using the authors' official
codes following the same settings in the original papers. We use the
results from FACT$^{\dagger}$ and CIRL$^{\dagger}$ for fair comparison
with our method. The best and second-best results are highlighted
in bold and underlined respectively.\label{tab:Leave-out-domain-out-results-PACS-OfficeHome-Digits}}
\end{table*}

Following previous works~\cite{lv2022causality,mahajan2021domain,zhou2020learning},
we evaluate our proposed methods on three standard OOD generalisation
benchmark datasets described below.

\textbf{PACS}~\cite{li2017deeper} consists of 9991 images from 4
domains namely Photo (\emph{P}), Art-painting (\emph{A}), Cartoon
(\emph{C}), Sketch (\emph{S}). These domains have largely different
styles, making this dataset challenging for OOD generalisation. There
are 7 classes in each domain. We use the training-validation-test
split in~\cite{li2017deeper} for fair comparison.

\textbf{Office-Home}~\cite{venkateswara2017deep} contains 15,500
images of office and home objects divided into 65 categories. The
images are from 4 domains: Art (\emph{A}), Clipart (\emph{C}), Product
(\emph{P}), and Real-world (\emph{R}). We use the training-validation-test
split in~\cite{zhou2021mixstyle}.

\textbf{Digits-DG}~\cite{zhou2020deep} is comprised of 4 digit domains
namely MNIST~\cite{lecun1998gradient} (\emph{MN}), MNIST-M~\cite{ganin2015unsupervised}
(\emph{MM}), SVHN~\cite{netzer2011reading} (\emph{SV}) and SYN~\cite{ganin2015unsupervised}
(\emph{SY}) with 10 digit classes (0-9) in each domain. These domains
are very different in font style, stroke colour, and background. We
follow~\cite{zhou2020deep,zhou2021domain} to select 600 images of
each class from each domain, where 80\% and 20\% of the selected data
are used for training and validation respectively.

All images have the size of $224\times224$ for PACS and Office-Home,
and $32\times32$ for Digits-DG.

\subsection{Experimental Setting}

We use the leave-one-domain-out strategy~\cite{li2017deeper,zhou2021mixstyle}
for evaluation: a model is tested on an unseen domain after training
on all the remaining ones.

\textbf{Classifier details}: Following~\cite{xu2021fourier,zhou2021mixstyle,li2021confounder},
we use pre-trained ResNet-18~\cite{he2016deep,deng2009imagenet}
backbone for PACS, Office-Home, and a small convolutional network~\cite{zhou2020deep}
trained from scratch for Digits-DG. For all datasets, we train the
network $F$ for $50$ epochs using an SGD optimiser with momentum
$0.9$, weight decay $5\times10^{-4}$, batch sizes $64$ for Digits-DG
and $32$ for PACS and Office-Home. For Digits-DG, the initial learning
rate is $0.03$ and decayed by $0.1$ after every $20$ epochs, while
those for PACS and Office-Home are initialised at $0.001$ and follow
the cosine annealing schedule. We deploy the standard augmentation
protocol in~\cite{zhou2021mixstyle,li2021confounder}, where only
random translation is used for Digits-DG while random flipping, random
translation, color-jittering, and random gray-scaling are used for
PACS and Office-Home.

\textbf{NST model details}: For each dataset, we fine-tune the decoder
of the pre-trained AdaIN NST~\cite{huang2017arbitrary} model for
$70$ epochs with all hyper-parameters referenced from~\cite{huang2017arbitrary},
whereas its encoder is a \emph{fixed} pre-trained VGG-19~\cite{simonyan2014very}
from ImageNet. Since NST models do not work well on small-size images,
for Digits-DG, we upscale images to the size of $(224\times224)$
before downscaling to $(32\times32)$ for the classifier $F$. We
also adopt the leave-one-domain-out strategy when training the NST
model and then use the corresponding version for classification.

We find the hyper-parameters $\alpha$ and $\beta$ of $\Model$ accordingly
for each domain based on a grid search algorithm on the validation
set, whereas the hyper-parameters of the FST in $\ModelFourier$ are
referenced from~\cite{lv2022causality,xu2021fourier}. In both the
training and inference phases, for each method, we sample $K=3$ training
samples from \emph{different domains }and compute $P(Y|\do(X))$ using
the same formula in Eqs.~\ref{eq:final_loss-FAST},~\ref{eq:final_loss-FAFT},
and~\ref{eq:final_loss-FAGT}. More implementation details are given
in the supplementary. We repeat the experiment ten times with different
seed values and report the average results.

\subsection{Experimental Results}

In Table~\ref{tab:Leave-out-domain-out-results-PACS-OfficeHome-Digits},
we show the domain generalisation results on the three benchmark datasets.
The baselines are divided into two groups: non-causality-based methods
(from DeepAll~\cite{zhou2020deep} to FACT~\cite{xu2021fourier}),
and causality-based methods (from MatchDG~\cite{mahajan2021domain}
to CIRL~\cite{lv2022causality}).

It is clear that our methods significantly outperform almost all non-causality-based
baselines on all datasets, suggesting the advantage of causal inference
for OOD generalisation. For example, $\ModelGeneral$ achieves about
4.5/0.9\% and 1.6/3.3\% higher accuracy than MixStyle~\cite{zhou2021mixstyle}
and RSC~\cite{huang2020self} respectively on PACS/Office-Home. It
is worth noting that MixStyle and RSC are both quite strong baselines.
However, they still rely on the superficial statistical dependence
$P(Y|X)$, instead of our modelling the causal dependence $P\left(Y|\do(X)\right)$.
Our methods also improve by about 4\% and 2\% over L2A-OT~\cite{zhou2020learning}
and FACT~\cite{xu2021fourier} accordingly on PACS. However, $\ModelGeneral$
achieves slightly lower accuracies compared to L2A-OT in some domains
of Office-Home, while beating it by a large margin of about 5\% on
the most challenging domain \emph{Clipart. }On Office-Home and Digits-DG,
although $\ModelGeneral$ remains competitive with FACT on almost
all domains, our performance drops a little in \emph{Real-world} of
Office-Home and \emph{SVHN} of Digits-DG. We note that although we
try but cannot reproduce the results of FACT reported in their paper.
In the source codes of FACT, the authors do not provide the config
files for Office-Home and Digits-DG datasets. Thus, we follow the
experimental setting and hyper-parameters shown in their paper and
rerun their source codes ten times but cannot replicate the reported
results. We use the best results of FACT among our reruns for fairness
and report as FACT$^{\dagger}$. Our methods show superior performance
over FACT$^{\dagger}$ in all datasets.

It can be observed that $\Model$ surpasses MatchDG~\cite{mahajan2021domain},
a quite state-of-the-art (SOTA) causality-based method that aims to
learn domain-invariant representation, by about 5.0\% and 1.3\% in
\emph{Art-Painting} and \emph{Sketch} on PACS respectively. We compare
our methods with CICF~\cite{li2021confounder}, a recent method that
interprets the causal front-door formula (Eq.~\ref{eq:frontdoor_adj_final})
via the meta-learning perspective. In some challenging domains (e.g.,\emph{
Clipart }of Office-Home or \emph{SYN }of Digits-DG) where testing
images appear with diverse domain-specific confounders such as colourful
backgrounds, our methods significantly outperform CICF. This suggests
that the style-based information of our style transfer algorithms
is more crucial to mitigating the spurious correlations of confounders
than the gradient-based meta-learning information of CICF. On PACS,
$\ModelGeneral$ beats CIRL in terms of the average accuracy over
all domains. It is worth noting that CIRL is a current SOTA method
in OOD generalisation and an improvement of 1.4\% over it in \emph{Art-painting
}of PACS is considerable. However, on Office-Home and Digits-DG, CIRL
outperforms our methods in some domains (e.g., \emph{Art} of Office-Home,
\emph{MNIST-M} and \emph{SVHN} of Digits-DG). The improvement of CIRL
is owing to their Factorisation and Adversarial Mask modules, making
the method more complicated than ours. Besides the causal intervention,
CIRL also makes use of an adversarial mask that significantly improves
its performance. From Table 5 in the CIRL paper, it is clear that
if the adversarial mask is removed (Variant 3), the performance of
CIRL on PACS will drop by nearly 1\% to 85.43 and is much worse than
our proposed method. By contrast, our methods learn a plain classifier
with only the maximum interventional likelihood objective and can
achieve more than 86\% average accuracy on PACS. Similar to FACT,
we try but cannot reproduce the results of CIRL. We rerun their source
codes ten times, and report the best results among our reruns as CIRL$^{\dagger}$
for fairness. Compared to CIRL$^{\dagger}$, $\ModelGeneral$ surpasses
it by 1.1\%, 0.7\%, and 3.2\% on PACS, Office-Home, and Digits-DG
respectively.

\paragraph*{Discussion about why our method works}

\begin{figure}
\begin{centering}
\includegraphics[width=1\columnwidth]{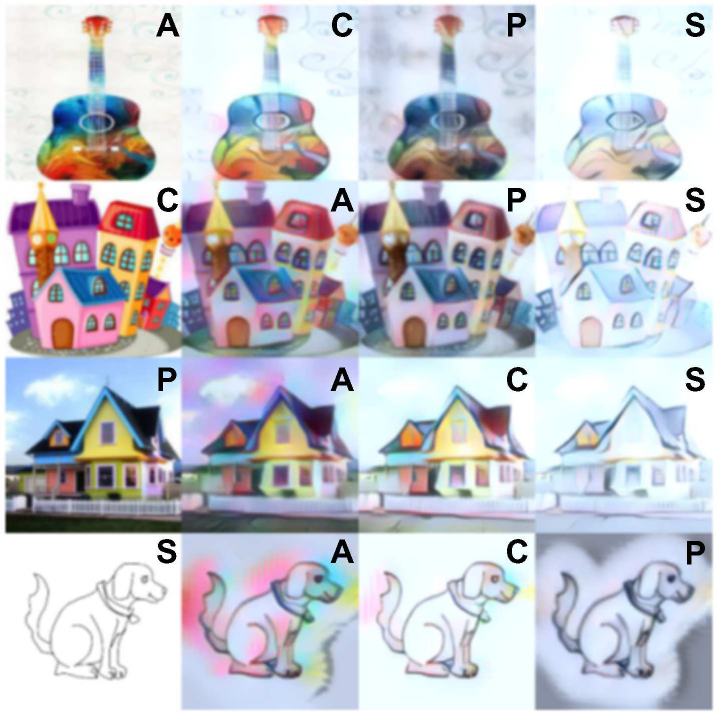}
\par\end{centering}
\caption{Visualisation of how unseen novel domains can be translated back to
the learnt ones without hurting the main content by \emph{$\protect\Model$}
on PACS. Each testing image is plotted in the first column, where
the remaining columns visualise its translated version in the trained
domains. Domain labels are noted at the upper right corner of each
image. \label{fig:Visualizations-of-how-mapping-FAST}}
\end{figure}

We argue three main reasons behind the success of our methods in OOD
generalisation. Owing to the AdaIn NST model, our methods could map
any novel domain back to the learnt ones without hurting the content
of images, which is beneficial for computing $P(Y|do(X))$. We visually
justify this point in Fig.~\ref{fig:Visualizations-of-how-mapping-FAST}. 

On the other hand, our great performance would be largely contributed
by modelling the causal dependence via front-door adjustment, which
we show to be invariant across domains in Sections~\ref{sec:Problem_Formulation},~\ref{sec:Method}.
In Table~\ref{tab:Comparison-causal-with-non-causal-1-1}, we show
the effectiveness of using $P(Y|do(X))$ over $P(Y|X)$ in both the
training and inference phases, where the method with causality in
both phases clearly outperforms the others. Many recent state-of-the-art
causality-based methods in OOD generalisation enforce regularisation
with different data transformations on the learnt feature to achieve
the domain-invariant feature~\cite{mahajan2021domain,wang2022out,lv2022causality}.
Although achieving promising results, the regularisation objectives
may have gradient conflicts with other learning objectives of the
model, leading to inferior performance in some cases~\cite{chen2023pareto}.

Asides, our front-door adjustment solution is orthogonal and more
flexible, since various stylised images can be taken into account
to eliminate spurious correlations of confounders, favouring the causal
dependence $P(Y|do(X)).$ This is validated by the superior performance
of $\ModelGeneral$ across all the experimented datasets. Further,
our estimation of the causal front-door formula via style transfer
algorithms in Eq.~\ref{eq:frontdoor_adj_final} is reasonable and
general, enabling various stylised images in both $\Model$ and $\ModelFourier$
to be taken into account for mitigating the negative correlations
of confounders. 

\begin{table}
\begin{centering}
\begin{tabular}{|c|c|c|c|c|c|c|}
\hline 
\multicolumn{2}{|c|}{Causal Dependence} & \multicolumn{5}{c|}{PACS}\tabularnewline
\hline 
Training & Inference & A & C & P & S & \emph{Avg.}\tabularnewline
\hline 
\hline 
- & - & 84.8 & 78.8 & 96.2 & 79.1 & 84.8\tabularnewline
\hline 
\Checkmark{} & - & 85.1 & 79.1 & 96.4 & 79.0 & 84.9\tabularnewline
\hline 
\Checkmark{} & \Checkmark{} & \textbf{87.5} & \textbf{80.9} & \textbf{96.9} & \textbf{81.9} & \textbf{86.8}\tabularnewline
\hline 
\end{tabular}
\par\end{centering}
\caption{The performance of our $\protect\ModelGeneral$ method with and without
using the causal dependence in the training and inference phases.
All methods are conducted on the same experimental setting, the method
without causality in both the phases uses augmented samples $\tilde{x}$
or $\hat{x}$ for a fair comparison. Overall, the method with causality
in both phases clearly outperforms the others.~\label{tab:Comparison-causal-with-non-causal-1-1}}
\end{table}

\begin{figure}
\begin{centering}
\includegraphics[width=0.96\columnwidth]{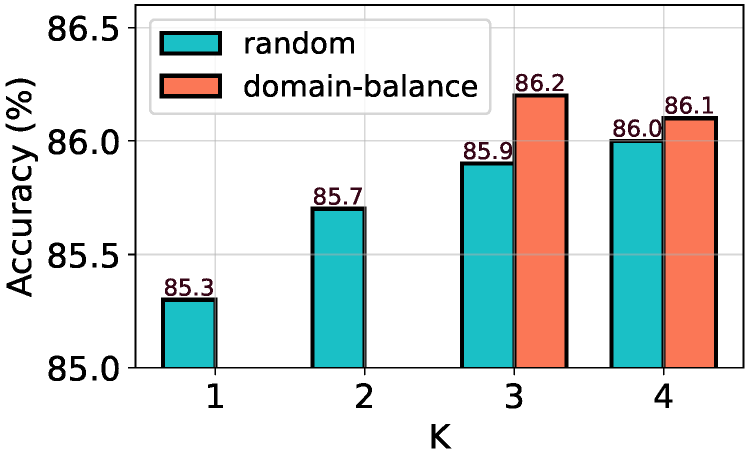}
\par\end{centering}
\caption{Accuracies (in \%) of $\protect\Model$ on PACS with different numbers
of style images $K$. ``random'' and ``domain-balance'' denote
two sampling strategies of style images. \label{fig:ablation-k}}
\end{figure}

\begin{figure*}[t]
\begin{centering}
\subfloat[Photo]{\begin{centering}
\includegraphics[width=0.24\textwidth]{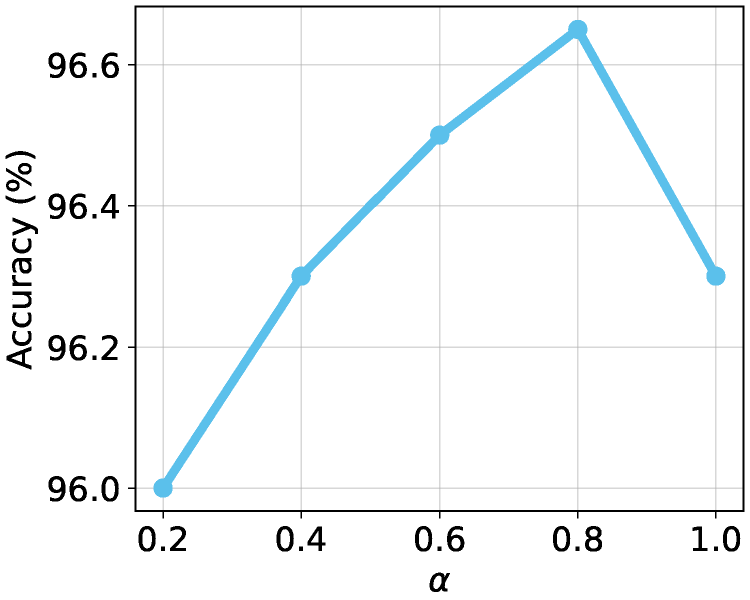}
\par\end{centering}
}\subfloat[Art-painting]{\begin{centering}
\includegraphics[width=0.24\textwidth]{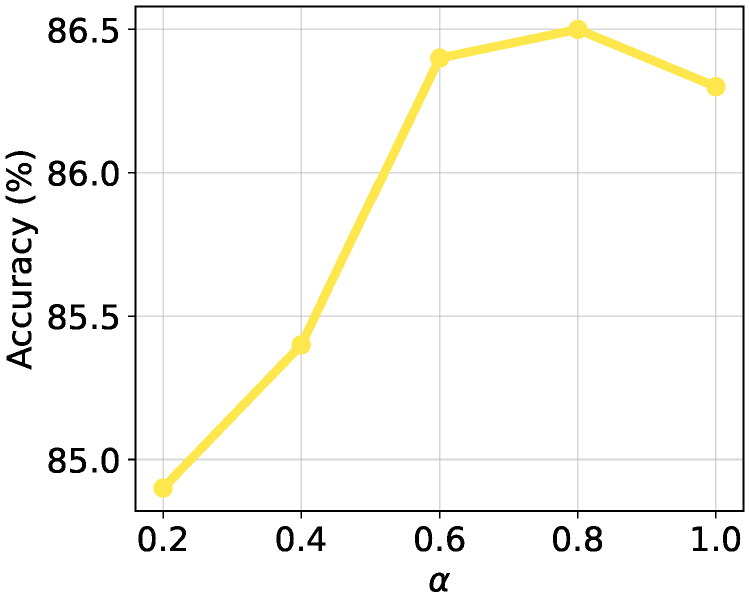}
\par\end{centering}
}\subfloat[Cartoon]{\begin{centering}
\includegraphics[width=0.24\textwidth]{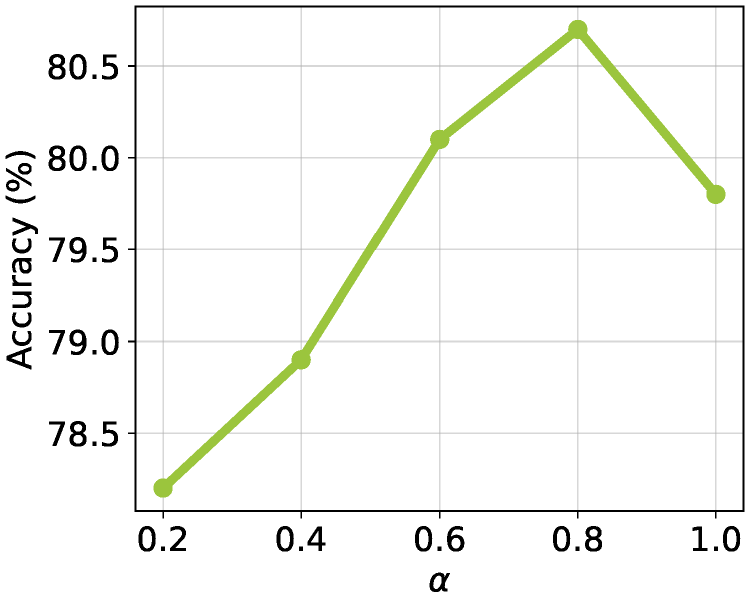}
\par\end{centering}
}\subfloat[Sketch]{\begin{centering}
\includegraphics[width=0.24\textwidth]{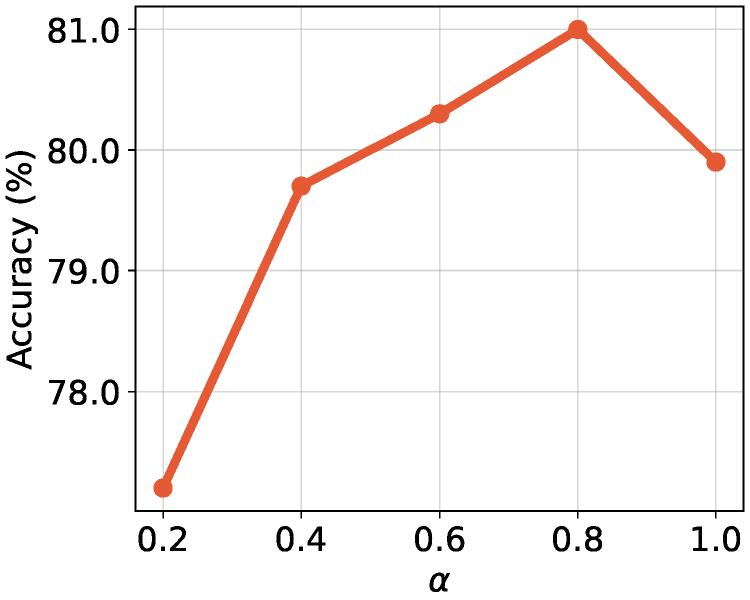}
\par\end{centering}
}
\par\end{centering}
\caption{The performance of $\protect\Model$ on the test domains of PACS with
different values of the style mixing coefficient $\alpha$.\label{fig:ablation-alpha}}
\end{figure*}

\subsection{Ablation Studies}

\begin{figure}[H]
\begin{centering}
\includegraphics[width=1\columnwidth]{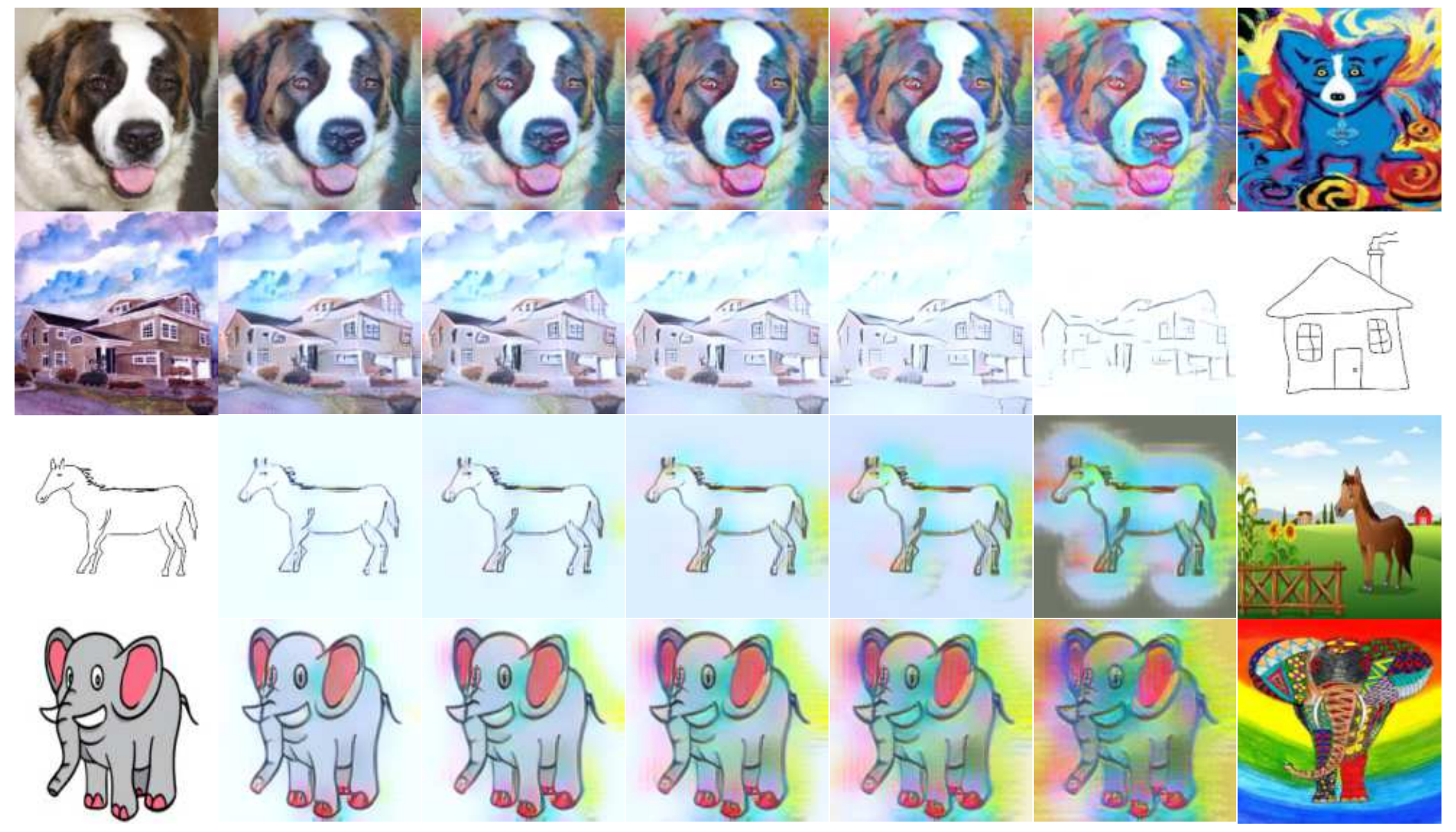}
\par\end{centering}
\caption{Visualisation of stylised images with different values of $\alpha$.
The first and last columns contain content and style images respectively.
The middle columns contain stylised images w.r.t. $\alpha$ being
0.2, 0.4, 0.6, 0.8, and 1.0 respectively.\label{fig:visualize-stylized-images}}
\end{figure}

\paragraph*{Number of style images}

We study the impact of the number of style images $K$ (Eq.~\ref{eq:final_loss-FAST})
on the performance of $\Model$. In Fig.~\ref{fig:ablation-k}, we
show the average results of $\Model$ on PACS with K varying from
1 to 4. We see that i) increasing the number of style images often
lead to better results when these images are sampled randomly from
the training data, and ii) we can improve the results further by ensuring
a balance in the number of sampled style images across domains.

\paragraph*{Trade-off between content and style}

In Fig.~\ref{fig:ablation-alpha}, we show the results of $\Model$
on each test domain of PACS with the style mixing coefficient $\alpha$
(Eq.~\ref{eq:stylized_z_itpl}) taking different values in \{0.2,
0.4, 0.6, 0.8, 1.0\}. It is clear that $\Model$ achieves higher accuracies
on all the test domains when $\alpha$ becomes larger (up to 0.8),
or equivalently, when $\tilde{x}$ more likely contains the style
of $x'$. This empirically justifies the reasonability of using style
transfer to model front-door adjustment. However, if $\alpha$ is
too large (e.g., $\alpha$ = 1.0), the performance of $\Model$ drops.
This is because too large $\alpha$ leads to the loss of content information
in the stylised images as shown in Fig.~\ref{fig:visualize-stylized-images}.

\section{Conclusion\label{sec:Conclusion}}

This paper studies the problem of Out-of-distribution (OOD) generalisation
from the causality perspective. We show that statistical learning
methods are ineffective for OOD generalisation due to spurious correlations
from domain-specific confounders into model training, resulting in
instability to domain shifting. The most commonly used causal inference
method, back-door adjustment, is unfeasible since the method requires
the observation of confounders, which is impractical in the setting.
We address the issue via front-door adjustment where features essential
to label prediction are considered the mediator variable. It is not
trivial to interpret the front-door formula, and we propose novel
methods that estimate it based on many style transfer models. We evaluated
our methods on several benchmark datasets. Experimental results show
the superiority of our method over existing baselines in OOD generalisation.

\balance

\bibliographystyle{ACM-Reference-Format}
\bibliography{reference}

\clearpage{}

\appendix
\global\long\def\do{\mathrm{do}}%

In this supplementary material, we first provide a detailed derivation
of the front-door formula used in our method, and a proof sketch on
the fulfillment of our proposed algorithms with the front-door criterion
in Section~\ref{sec:Derivation-of-Front-door}. In Section~\ref{sec:Derivation-of-MIL-causal},
we give a complete derivation of the Maximum Interventional Log-likelihood
in our formulation of OOD generalisation under the causal perspective.
We show our experimental results with ResNet-50~\cite{he2016deep}
on PACS and Offfice-Home in Section~\ref{sec:More-experimental-results}.
A comprehensive description of experimental setup including parameter
settings is given in Section~\ref{sec:More-Details-on-experimental-settings}.
We make more ablation studies and discussion about our method in Section~\ref{sec:More-ablation-studies}.
Finally, we visualise several stylised images produced by our algorithms
in Section~\ref{sec:More-Visualisations}.

\section{Details on Front-door Adjustment in OOD Generalisation\label{sec:Derivation-of-Front-door}}

\subsection{Derivation of Front-door Adjustment\label{subsec:Derivation-of-Front-door-detailed}}

In this section, we provide a mathematical derivation for the front-door
adjustment of $P\left(Y|\do(x)\right)$ in the main text. Recalling
that for front-door adjustment, we use a mediator variable $Z$ between
$X$ and $Y$ which satisfies the following conditions: (i) All directed
paths from $X$ to $Y$ flow through $Z$. (ii) $X$ blocks all back-door
paths from $Z$ to $Y$. (iii) There are no unblocked back-door paths
from $X$ to $Z$. Then, we can identify $P\left(Y|\do(x)\right)$
as below:

\begin{align}
P\left(Y|\do(x)\right) & =\sum_{z}p(z|\do(x))P(Y|z,\do(x))\label{eq:frontdoor_1}\\
 & =\sum_{z}p(z|x)P(Y|\do(z),\do(x))\label{eq:frontdoor_2}\\
 & =\sum_{z}p(z|x)P(Y|\do(z))\label{eq:frontdoor_3}\\
 & =\sum_{z}p(z|x)\left(\sum_{x'}p(x')P(Y|z,x')\right)\label{eq:frontdoor_4}\\
 & =\mathbb{E}_{p(z|x)}\mathbb{E}_{p(x')}[P(Y|z,x')]\label{eq:frontdoor_5}
\end{align}
Eq.~(\ref{eq:frontdoor_1}) follows the standard probabilistic rules
to incorporate $z$ into the formula of $P(Y|\do(x))$. In Eq.~(\ref{eq:frontdoor_2}),
$p(z|x)=p(z|\do(x))$ due to condition (iii); $P(Y|z,\do(x))=p(Y|\do(z),\do(x))$
due to condition (ii). In Eq.~(\ref{eq:frontdoor_3}), $P(Y|\do(z),\do(x))=P(Y|\do(z))$
due to condition (i). In Eq.~(\ref{eq:frontdoor_4}), $\sum_{x'}p(x')P(Y|z,x')$
is just the back-door adjustment of $P(Y|\do(z))$ with condition
(ii) satisfied. Eq.~(\ref{eq:frontdoor_5}) is a rewriting of Eq.~(\ref{eq:frontdoor_4})
using the expectation notation.

\subsection{Satisfaction of the front-door criterion~\label{subsec:fulfillment-of-front-door}}

In this section, we provide a proof sketch that $Z$ as semantic features
in $X$ will satisfy the front-door criterion.

\textbf{Claims}: Let $Z$ is a variable that captures all and only
the semantics in $X$. Then, $Z$ will satisfy the front-door criterion.

\textbf{Justification}: 
\begin{itemize}
\item \emph{All the directed paths $X\rightarrow Y$ are intercepted by
$Z$}. Intuitively, the label variable $Y$ is defined from $X$ only
by its semantic features that are \emph{always} predictive of $Y$.
In reality, the annotators might only use these features to define
$Y$ from observing $X$ (e.g., by noticing main objects in $X$).
In our methods, $Z$ learnt by $\Model$ and $\ModelFourier$ fulfills
the above condition. In $\ModelFourier$, $Z$ is regarded as the
phase component of the Fourier Transformation of $X$, which has been
shown to preserve most of the high-level semantics such as edges or
contours of objects~\cite{xu2021fourier}. According to~\cite{huang2017arbitrary},
once the AdaIN NST model has been trained well, its output $\tilde{z}$
captures and preserves the semantics of $x$, which are transferred
from $z$.
\item \emph{No unblocked back-door paths from $X$ to $Z$. }This requirement
suggests that we could estimate the causal relationship $(X\rightarrow Z)$
by observing training images, i.e., $P(Z|do(X))=P(Z|X).$ In both
$\Model$ and $\ModelFourier$, $Z$ is a \emph{mapping from $X$}
that sufficiently extract only semantic features in $X$ via the encoder
$E$ of the AdaIN NST model and the Fourier Transformation respectively.
It is worth noting that the trivial identity mapping should not be
appropriate, because it may map nuisance confouding features into
$Z$. Overall, $Z$ is defined only from $X,$ thus $Z$ has no causal
parent other than $X$ in $\mathcal{G}$ and there will be no confounder
between $X$ and $Z$. Therefore, we have $P(Z|X,d,u)=P(Z|X)$ for
any confounder $d,u$ and we can derive $P(Z|do(X))$ as follows: 
\end{itemize}
\begin{align}
P(Z|do(X)) & =\sum_{d,u}P(d,u)P(Z|d,u,X)\\
 & =\sum_{d,u}P(d,u)P(Z|X)\\
 & =P(Z|X)
\end{align}

\begin{itemize}
\item \emph{$X$ blocks all back-door paths from $Z$ to $Y$. }Given no
confounder that connects to $Z$, according to the causal graph shown
in Fig.~\ref{fig:method_overview}(b), we can see that there is only
one back-door path from $Z$ to $Y$, i.e., $Z\leftarrow X\leftarrow(D,U)\rightarrow Y$.
Therefore, we can condition on $X$ to block all the back-door paths
from $Z$ to $Y.$ 
\end{itemize}
Finally, we conclude that the variable $Z$ as all and only semantic
features in $X$ in our method will satisfy the front-door criterion.

\begin{table*}[t]
\begin{centering}
\begin{tabular}{c|cccc|c||cccc|c}
\hline 
\multirow{2}{*}{Methods} & \multicolumn{5}{c||}{PACS} & \multicolumn{5}{c}{Office-Home}\tabularnewline
\cline{2-11} \cline{3-11} \cline{4-11} \cline{5-11} \cline{6-11} \cline{7-11} \cline{8-11} \cline{9-11} \cline{10-11} \cline{11-11} 
 & A & C & P & S & \emph{Avg.} & A & C & P & R & \emph{Avg.}\tabularnewline
\hline 
\hline 
ERM~\cite{Vapnik1998} & 84.7 & 80.8 & 97.2 & 79.3 & 85.5 & 63.1 & 51.9 & 77.2 & 78.1 & 67.6\tabularnewline
\hline 
MMD-AAE~\cite{Li_2018_CVPR} & 86.1 & 79.4 & 96.6 & 76.5 & 84.7 & 60.4 & 53.3 & 74.3 & 77.4 & 66.4\tabularnewline
RSC~\cite{huang2020self} & 85.4 & 79.7 & 97.6 & 78.2 & 85.2 & 60.7 & 51.4 & 74.8 & 75.1 & 65.5\tabularnewline
MixStyle~\cite{zhou2021mixstyle} & 86.8 & 79.0 & 96.6 & 78.5 & 85.2 & - & - & - & - & -\tabularnewline
CCM~\cite{miao2022domain} & - & - & - & - & 87.0 & - & - & - & - & 69.7\tabularnewline
SAGM~\cite{wang2023sharpness} & 87.4 & 80.2 & \textbf{98.0} & 80.8 & 86.6 & 65.4 & 57.0 & \textbf{78.0} & \textbf{80.0} & 70.1\tabularnewline
FACT~\cite{xu2021fourier} & 89.6 & 81.8 & 96.8 & 84.5 & 88.2 & - & - & - & - & -\tabularnewline
\hline 
\hline 
MatchDG~\cite{mahajan2021domain} & 85.6 & 82.1 & \uline{97.9} & 78.8 & 86.1 & - & - & - & - & -\tabularnewline
CICF$^{\dagger}$~\cite{li2021confounder} & 89.7 & 82.2 & \uline{97.9} & \uline{86.2} & \uline{89.0} & 63.1 & \uline{59.4} & 77.4 & 78.1 & 69.5\tabularnewline
CIRL~\cite{lv2022causality} & \textbf{90.7} & \textbf{84.3} & 97.8 & \textbf{87.7} & \textbf{90.1} & - & - & - & - & -\tabularnewline
\hline 
\hline 
$\Model$ (ours) & 89.2 & 83.4 & 97.3 & 82.5 & 88.1 & 65.7 & \uline{59.4} & 77.7 & 78.7 & \uline{70.4}\tabularnewline
$\ModelFourier$ (ours) & 89.4 & 83.3 & 97.5 & 82.3 & 88.1 & \uline{65.9} & 58.3 & \uline{77.8} & 78.9 & 70.2\tabularnewline
$\ModelGeneral$ (ours) & \uline{90.4} & \uline{83.8} & 97.7 & 83.1 & 88.8 & \textbf{66.3} & \textbf{59.5} & 77.6 & \uline{79.0} & \textbf{70.6}\tabularnewline
\hline 
\end{tabular}
\par\end{centering}
\caption{Leave-one-domain-out classification accuracies (in \%) on PACS, Office-Home
with ResNet-50. In ERM, we simply train the classifier on all domains
except the test one. CICF$^{\dagger}$as reported in~\cite{li2021confounder}
is the original CICF method with AutoAugment~\cite{cubuk2019auto}.
The best and second-best results are highlighted in bold and underlined
respectively.\label{tab:Leave-out-domain-out-results-PACS-OfficeHome-ResNet50}}
\end{table*}

\section{Derivation of the Maximum Interventional Log-likelihood\label{sec:Derivation-of-MIL-causal}}

In this section, we provide a detailed mathematical derivation of
the Maximum Interventional Log-likelihood (MIL) criterion presented
in Section~\ref{subsec:Causal-Inference-Perspective}. Assuming there
is a causal (interventional) distribution $P\left(Y|\do(X)\right)$
and a training set $\mathcal{D}$ contains $n$ samples $\left\{ (x_{i},y_{i})\right\} _{i=1}^{n}$.
We consider $\do(x_{1})$, ..., $\do(x_{n})$ as $n$ interventions
on $X$ which lead to $n$ random potential outcomes $Y_{1}$, ...,
$Y_{n}$. The joint probability of these potential outcomes is $P\left(Y_{1},...,Y_{n}|\do\left(x_{1}\right),...,\do\left(x_{n}\right)\right)$.
Since modelling this distribution is difficult and impractical, we
factorise it as follows: 
\begin{align}
 & P\left(Y_{1},...,Y_{n}|\do\left(x_{1}\right),...,\do\left(x_{n}\right)\right)\nonumber \\
 & =\prod_{i=1}^{n}P\left(Y_{i}|\do\left(x_{1}\right),...,\do\left(x_{n}\right)\right)\label{eq:Factorization_1}\\
 & =\prod_{i=1}^{n}P\left(Y_{i}|\do\left(x_{i}\right)\right)\label{eq:Factorization_2}
\end{align}

Eq.~(\ref{eq:Factorization_1}) depends on a condition saying that
the outcome of a particular item $i$ is independent of the outcomes
of other items once the interventions of all the items are provided.
This condition generally satisfies since we usually consider causal
graphs containing no loop on $Y$. Eq.~(\ref{eq:Factorization_2})
is based on the \emph{``no-interference''} assumption~\cite{yao2021survey}
saying that the potential outcome $Y_{i}$ of an item $i$ does not
depend on the interventions $\do(x_{j})$ of other items $j\neq i$.
The factorised distribution in Eq.~(\ref{eq:Factorization_2}) allows
us to model $P\left(Y|\do(X)\right)$.

Next, we want to know the most probable potential outcome under the
intervention $\do\left(x_{i}\right)$ given that the outcome $y_{i}$
w.r.t. the treatment $x_{i}$ was observed. The \emph{``consistency''}
assumption~\cite{yao2021survey} provides us with an answer to this
question. It states that the observed outcome $y_{i}$ w.r.t. to a
treatment $x_{i}$ of an item $i$ is the potential outcome $Y_{i}$
under the intervention $\do\left(x_{i}\right)$. This means $Y_{i}|\do\left(x_{i}\right)$
is likely to be $y_{i}$, and it is reasonable to maximise $P\left(Y_{i}=y_{i}|\do(x_{i})\right)$.

Therefore, given the ``no-interference'' and ``consistency'' assumptions,
we can learn a causal model $P_{\theta}\left(Y|\do(X)\right)$ from
an observational training set $\mathcal{D}$ by optimising the following
objective: 
\[
\theta^{*}=\argmax{\theta}\mathbb{E}_{(x,y)\sim\mathcal{D}}\left[\log P_{\theta}\left(Y=y|\do(x)\right)\right]
\]
which we refer to as the \emph{Maximum Interventional Log-likelihood
(MIL)}.

\section{Results with ResNet-50 on PACS and Office-Home\label{sec:More-experimental-results}}

We conducted more experiments on PACS and Office-Home with backbone
ResNet-50. All hyper-parameters are the same as those used with ResNet-18.
In Table~\ref{tab:Leave-out-domain-out-results-PACS-OfficeHome-ResNet50},
we show the domain generalisation results on PACS and Office-Home
with ResNet-50. The baselines are divided into two groups: non-causality-based
methods (from MMD-AAE~\cite{Li_2018_CVPR} to FACT~\cite{xu2021fourier}),
and causality-based methods (from MatchDG~\cite{mahajan2021domain}
to CIRL~\cite{lv2022causality}). 

As can be seen from Table~\ref{tab:Leave-out-domain-out-results-PACS-OfficeHome-ResNet50},
our proposed methods outperform non-causality-based methods by a large
margin on both datasets, validating the effectiveness of causal inference
in OOD generalisation. For example, $\ModelGeneral$ achieves 3.6\%
and 0.6\% higher accuracy than MixStyle~\cite{zhou2021mixstyle}
and FACT~\cite{xu2021fourier} respectively on PACS. Among causality-based
methods, $\ModelGeneral$ remains competitive with CICF$^{\dagger}$~\cite{li2021confounder}
although our algorithm does not use as strong augmentation techniques
as CICF$^{\dagger}$. On the other hand, $\ModelGeneral$ surpasses
CICF$^{\dagger}$ by a large margin of 1.1\% on Office-Home, which
demonstrates our effectiveness in modelling front-door adjustment
via style transfer algorithms. As mentioned in Section~\ref{sec:Experiments},
the great performance of CIRL is owing to many sub-modules apart from
the causal intervention, such as the adversarial mask, while our methods
learn a plain classifier with only the maximum interventional likelihood
objective. This may be the reason why CIRL outperforms $\ModelGeneral$
in \emph{Sketch} on PACS, while our methods are competitive with CIRL
in other domains.

\section{Experimental Details\label{sec:More-Details-on-experimental-settings} }

We used the Pytorch Dassl toolbox~\cite{zhou2021domain} with Python
3.6 and Pytorch 1.10. All experiments were conducted on an Ubuntu
20.04 server with a 20-cores CPU, 384GB RAM, and Nvidia V100 32GB
GPUs. Apart from the hyper-parameters discussed in the main paper,
we also consider two other hyper-parameters including: i) the content-style
interpolation coefficient $\alpha$ in Eq.~(\ref{eq:stylized_z_itpl});
and ii) the coefficient $\beta$ which trades off between the prediction
of the original content image and those of multiple stylised images
in Eqs.~(\ref{eq:final_loss-FAST}),~(\ref{eq:final_loss-FAFT}),
and~(\ref{eq:final_loss-FAGT}). We used grid search on the validation
set to find optimal values for $\alpha$ and $\beta$, which are shown
in Table~\ref{tab:Hyper-parameters-chosen}.

\begin{table}
\begin{centering}
{\small{}}%
\begin{tabular}{c|c|c|c}
\hline 
\multirow{2}{*}{{\small{}Domains}} & \multicolumn{1}{c|}{{\small{}$\Model$}} & {\small{}$\ModelFourier$} & \multicolumn{1}{c}{{\small{}$\ModelGeneral$}}\tabularnewline
\cline{2-4} \cline{3-4} \cline{4-4} 
 & {\small{}$\alpha$, $\beta$} & {\small{}$\beta$} & {\small{}$\alpha$, $\beta$}\tabularnewline
\hline 
\hline 
{\small{}All} & {\small{}$\left[0.6,0.8\right]$,$\left[0.25,0.45\right]$} & {\small{}$\left[0.25,0.45\right]$} & {\small{}$\left[0.6,0.8\right],\left[0.15,0.3\right]$}\tabularnewline
\hline 
{\small{}A, P, R} & {\small{}$\left[0.4,0.6\right]$,$\left[0.35,0.55\right]$} & {\small{}$\left[0.35,0.55\right]$} & {\small{}$\left[0.4,0.6\right]$,$\left[0.15,0.3\right]$}\tabularnewline
{\small{}C} & {\small{}$\left[0.5,0.7\right]$,$\left[0.35,0.55\right]$} & {\small{}$\left[0.35,0.55\right]$} & {\small{}$\left[0.5,0.7\right]$,$\left[0.15,0.3\right]$}\tabularnewline
\hline 
{\small{}All} & {\small{}$\left[0.4,0.6\right]$,$\left[0.35,0.55\right]$} & {\small{}$\left[0.35,0.55\right]$} & {\small{}$\left[0.4,0.6\right]$,$\left[0.15,0.3\right]$}\tabularnewline
\hline 
\end{tabular}{\small\par}
\par\end{centering}
\caption{Details on hyper-parameters of our models (values of $\alpha$ in
Eq.~(\ref{eq:stylized_z_itpl}) and $\beta$ in Eqs.~\ref{eq:final_loss-FAST},~\ref{eq:final_loss-FAFT},
and~\ref{eq:final_loss-FAGT}) in each domain on PACS (\textbf{first}
row), Office-Home (\textbf{second} row), and Digits-DG (\textbf{third}
row). ``All'' stands for all domains. \label{tab:Hyper-parameters-chosen}}
\end{table}

\section{More ablation studies\label{sec:More-ablation-studies}}

\subsection{Number of style images}

\begin{figure}
\begin{centering}
\includegraphics[width=0.96\columnwidth]{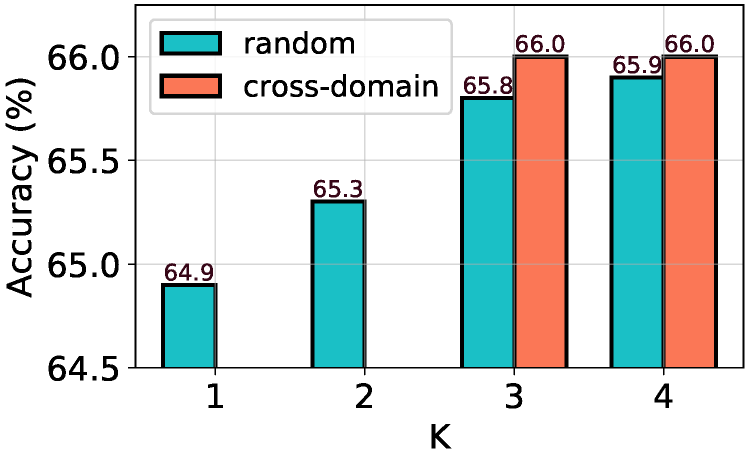}
\par\end{centering}
\caption{Accuracies (in \%) of $\protect\Model$ on Office-Home with different
numbers of style images $K$. ``random'' and ``domain-balance''
denote two sampling strategies of style images. \label{fig:ablation-k-office-home}}
\end{figure}

\begin{figure}
\begin{centering}
\includegraphics[width=0.96\columnwidth]{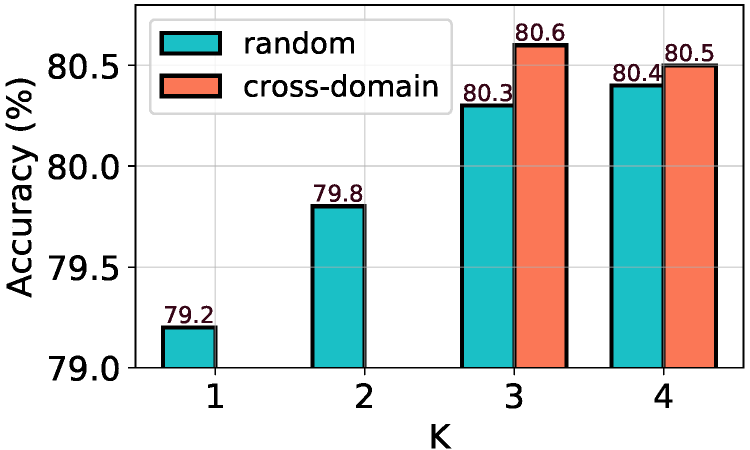}
\par\end{centering}
\caption{Accuracies (in \%) of $\protect\Model$ on Digits-DG with different
numbers of style images $K$. ``random'' and ``domain-balance''
denote two sampling strategies of style images. \label{fig:ablation-k-digits-dg}}
\end{figure}

We also report the performance of $\Model$ while varying the number
of style images $K$ (Eq.~\ref{eq:final_loss-FAST}) on Office-Home~\cite{venkateswara2017deep}
and Digits-DG~\cite{zhou2020deep} in Figures~\ref{fig:ablation-k-office-home}
and~\ref{fig:ablation-k-digits-dg}. Like the analysis on PACS~\cite{li2017deeper}
in the main paper, we observe that i) with the random sampling strategy,
the results are better with larger values of $K$, and ii) the results
are improved further when the balance in the number of sampled style
images across domains is established.

\subsection{Memory Usage and Running Time}

We compare our method $\Model$, a recent SOTA causality-based method
CIRL~\cite{lv2022causality}, and ERM~\cite{Vapnik1998} in terms
of the memory usage, total training time and testing time on PACS.
For a fair comparison, both methods use the same GPU resources and
experimental protocol (e.g., batch size) for both the training and
testing phases. Results are reported in Table~\ref{tab:Comparison-between-FAST-ERM-memory-time}.
Compared with ERM, $\Model$ introduces 2.3 GiB memory usage, about
more 4000 and 100 seconds in the training and testing phases respectively,
while gaining a significant improvement of 6.7\% on PACS. Although
our performance remains competitive with that of CIRL, our method
$\Model$ requires much less training and testing time than CIRL,
while the difference in allocated memory resources between the two
methods is meager.

\begin{table}
\begin{centering}
\begin{tabular}{|c|c|c|c|c|}
\hline 
 & Mem (GiB) & Training (s) & Testing (s) & Acc (\%)\tabularnewline
\hline 
\hline 
ERM~\cite{Vapnik1998} & 2.202 & 891 & 4 & 79.5\tabularnewline
\hline 
CIRL~\cite{lv2022causality} & 5.174 & 6381 & 243 & 86.3\tabularnewline
\hline 
FAST & 5.542 & 4832 & 107 & 86.2\tabularnewline
\hline 
\end{tabular}
\par\end{centering}
\caption{Comparison between our method $\protect\Model$, a recent SOTA causality-based
method CIRL~\cite{lv2022causality} and ERM~\cite{Vapnik1998} in
terms of the memory usage, total training time and testing time on
PACS. From left to right columns, we report the memory usage (GiB),
the training time (seconds), the testing time (seconds) and the average
accuracy (\%). The accuracy of baselines are referenced from \cite{lv2022causality}.
\label{tab:Comparison-between-FAST-ERM-memory-time}}
\end{table}

\section{Visualisations\label{sec:More-Visualisations}}

\subsection{Stylised images in Neural Style Transfer\label{subsec:Visualisation-of-stylized-FAST}}

\begin{figure*}
\begin{centering}
\includegraphics[height=1.05\textheight]{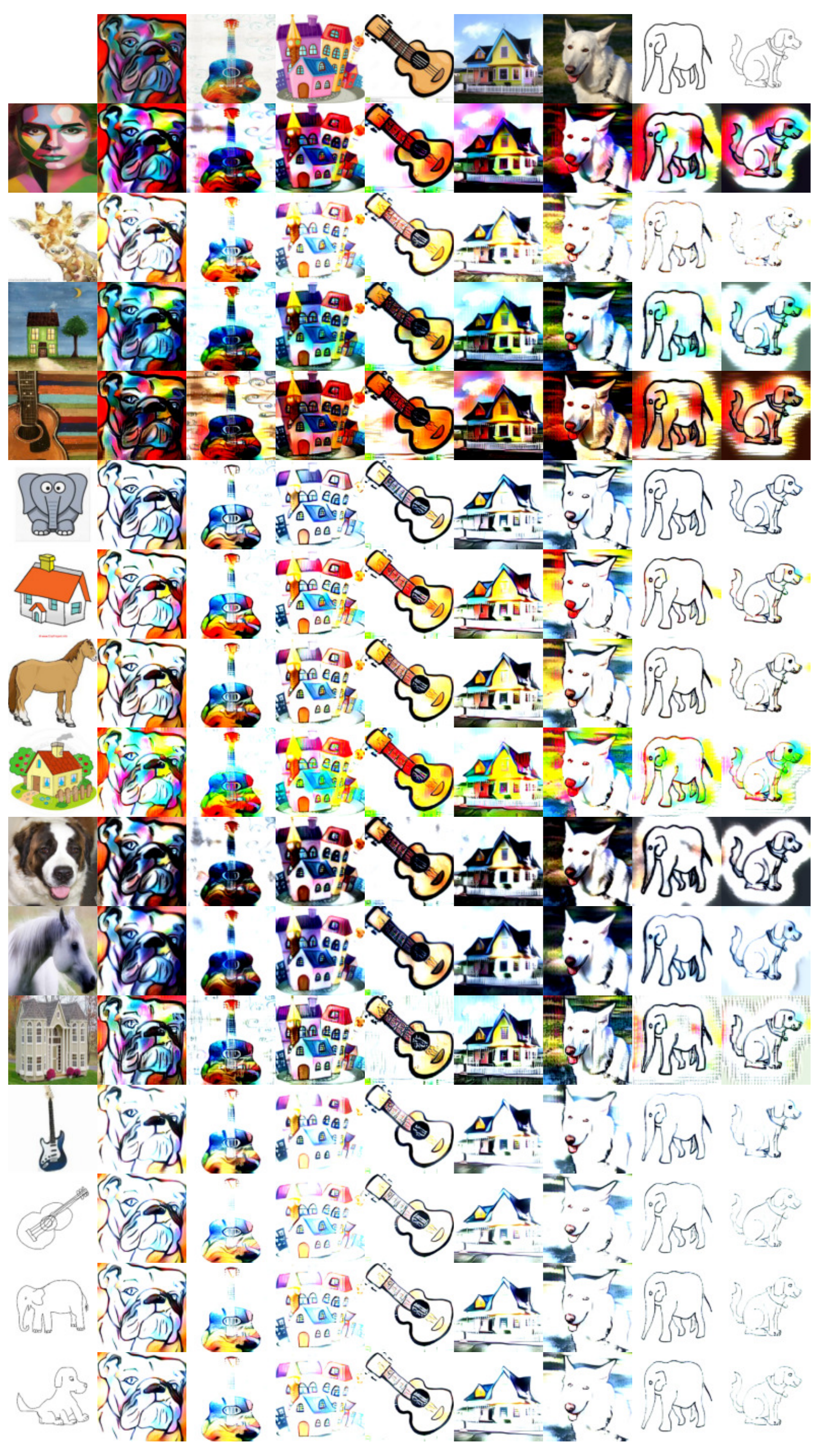}
\par\end{centering}
\caption{Stylised images produced by the AdaIN NST model of $\protect\Model$
on PACS with $\alpha=0.6$ in \emph{Sketch} and $\alpha=0.75$ in
the other domains. The content and style images are visualised on
the first row and column respectively. \label{fig:Visualisations-of-diverse-FAST-PACS}}
\end{figure*}

\begin{figure*}
\begin{centering}
\includegraphics[height=1.05\textheight]{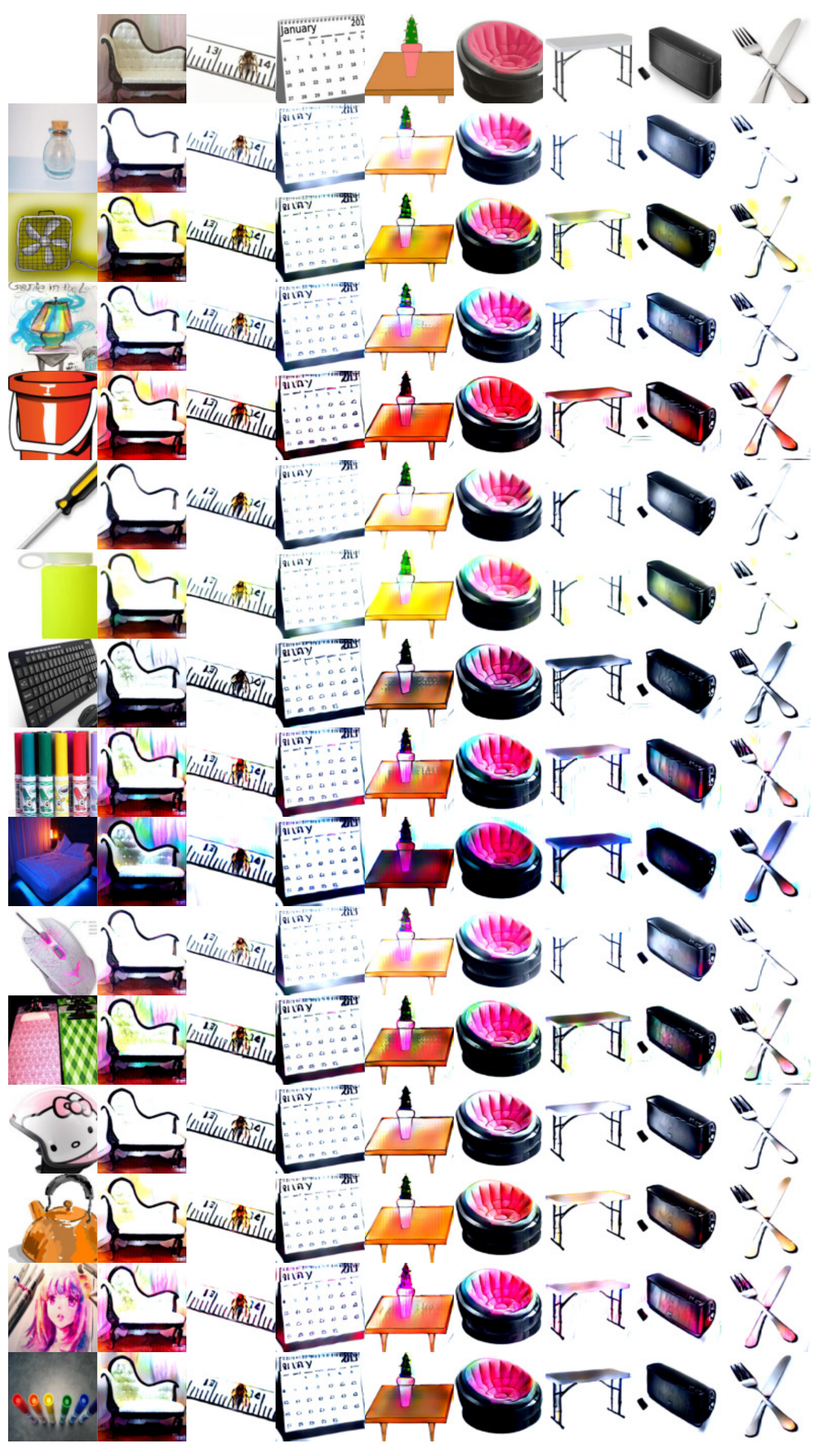}
\par\end{centering}
\caption{Stylised images produced by the AdaIN NST model of $\protect\Model$
on Office-Home with $\alpha=0.6$ in \emph{ClipArt} and $\alpha=0.45$
in the other domains. The content and style images are visualised
on the first row and column respectively.\label{fig:Visualisations-of-diverse-FAST-Office-Home}}
\end{figure*}

\begin{figure*}
\begin{centering}
\includegraphics[height=1\textheight]{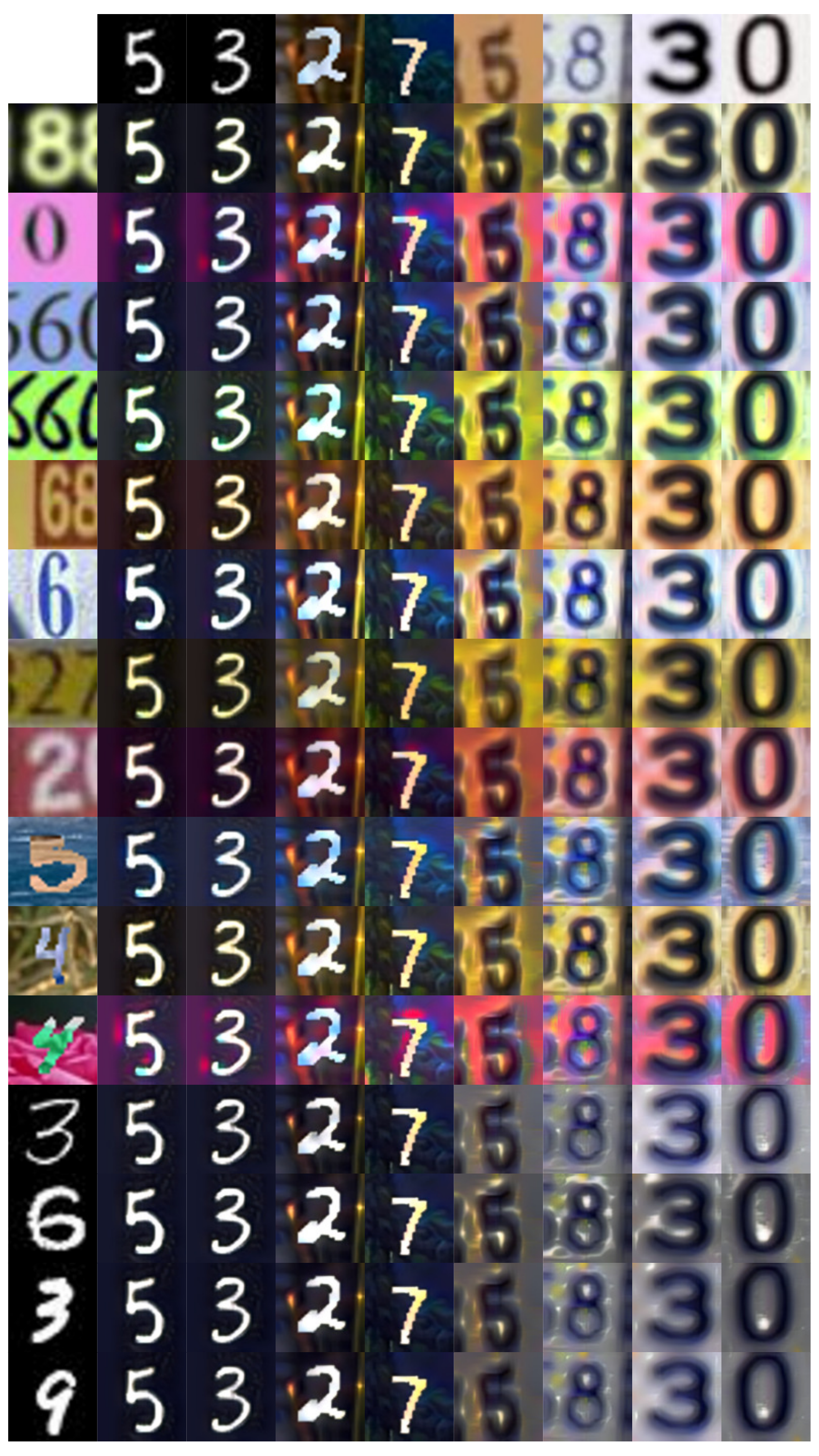}
\par\end{centering}
\caption{Stylised images produced by the AdaIN NST model of $\protect\Model$
on Digits-DG with $\alpha=0.45$ in all domains. The content and style
images are visualised on the first row and column respectively.\label{fig:Visualisations-of-diverse-FAST-digits-dg}}
\end{figure*}

We visualise stylised images produced by the AdaIN NST model of $\Model$
on PACS, Office-Home, and Digits-DG in Figures~\ref{fig:Visualisations-of-diverse-FAST-PACS},~\ref{fig:Visualisations-of-diverse-FAST-Office-Home},
and~\ref{fig:Visualisations-of-diverse-FAST-digits-dg} respectively.

\subsection{Stylised images in Fourier-based Style Transfer~\label{subsec:Visualisation-of-stylized-FAFT}}

\begin{figure*}
\begin{centering}
\includegraphics[height=1.05\textheight]{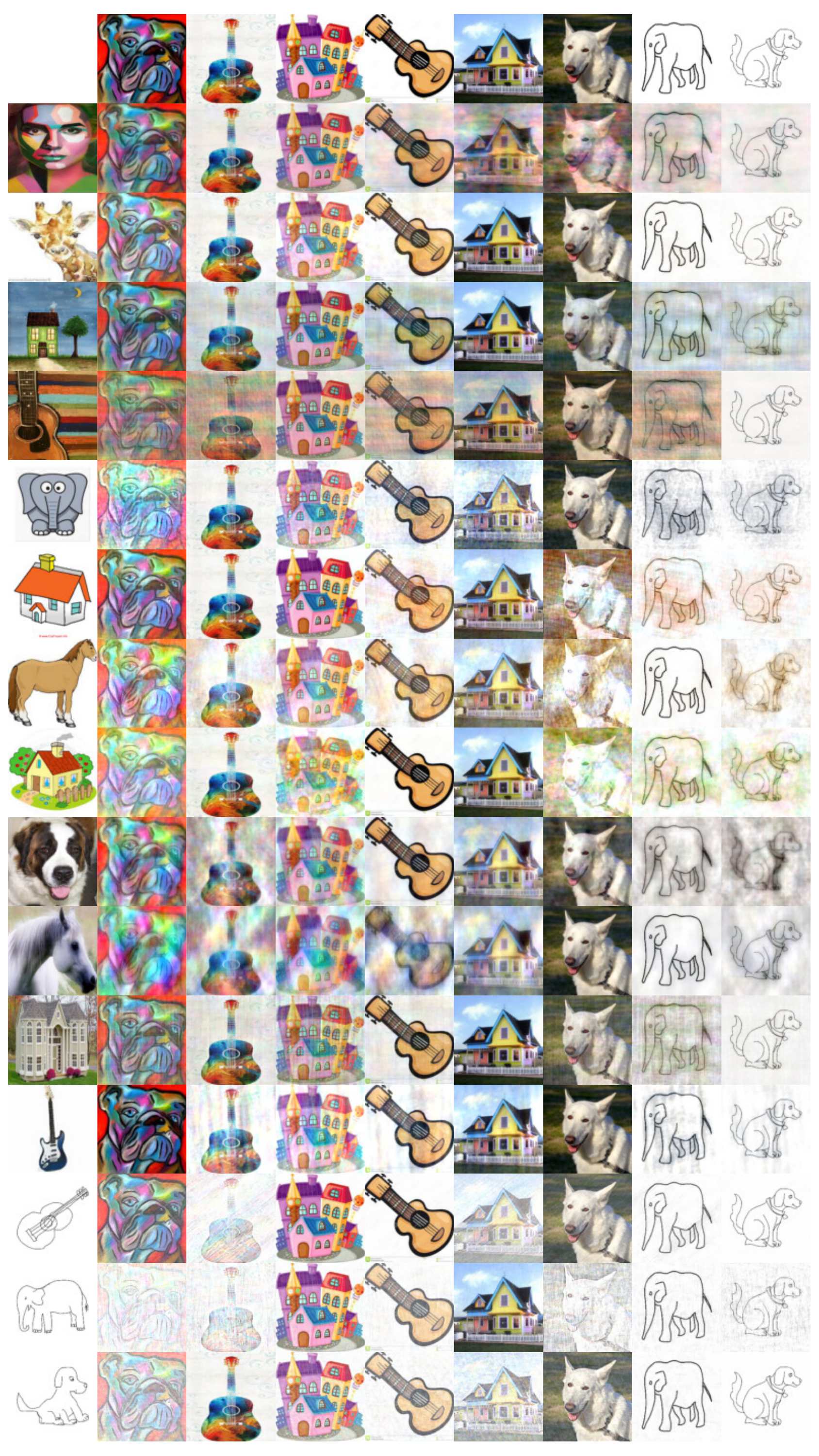}
\par\end{centering}
\caption{Stylised images produced by the Fourier-based Style Transfer of $\protect\ModelFourier$
on PACS. The content and style images are visualised on the first
row and column respectively. \label{fig:Visualisations-of-diverse-Fourier-PACS}}
\end{figure*}

\begin{figure*}
\begin{centering}
\includegraphics[height=1.05\textheight]{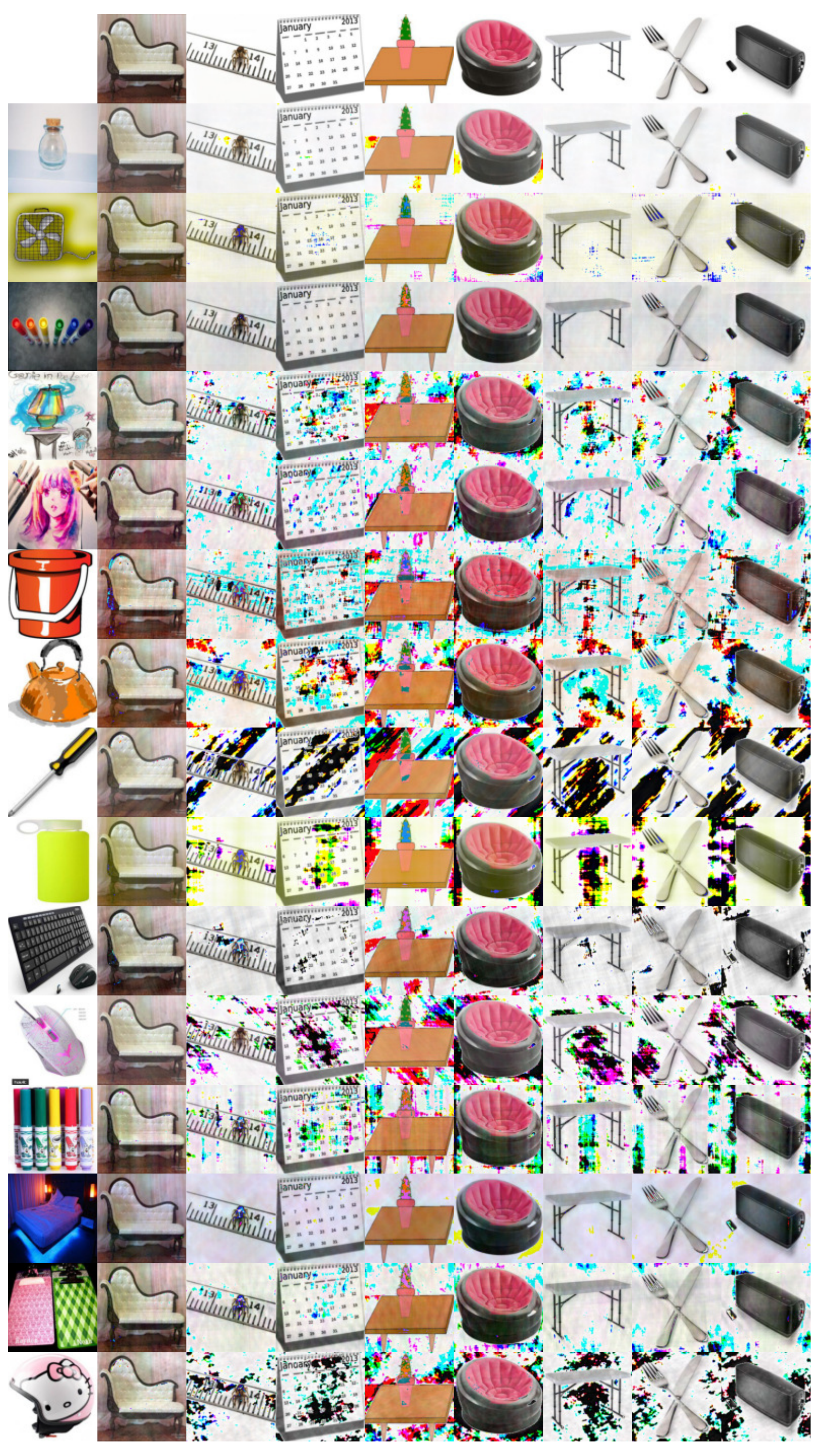}
\par\end{centering}
\caption{Stylised images produced by the Fourier-based Style Transfer of $\protect\ModelFourier$
on Office-Home. The content and style images are visualised on the
first row and column respectively.\label{fig:Visualisations-of-diverse-Fourier-Office-Home}}
\end{figure*}

\begin{figure*}
\begin{centering}
\includegraphics[height=1\textheight]{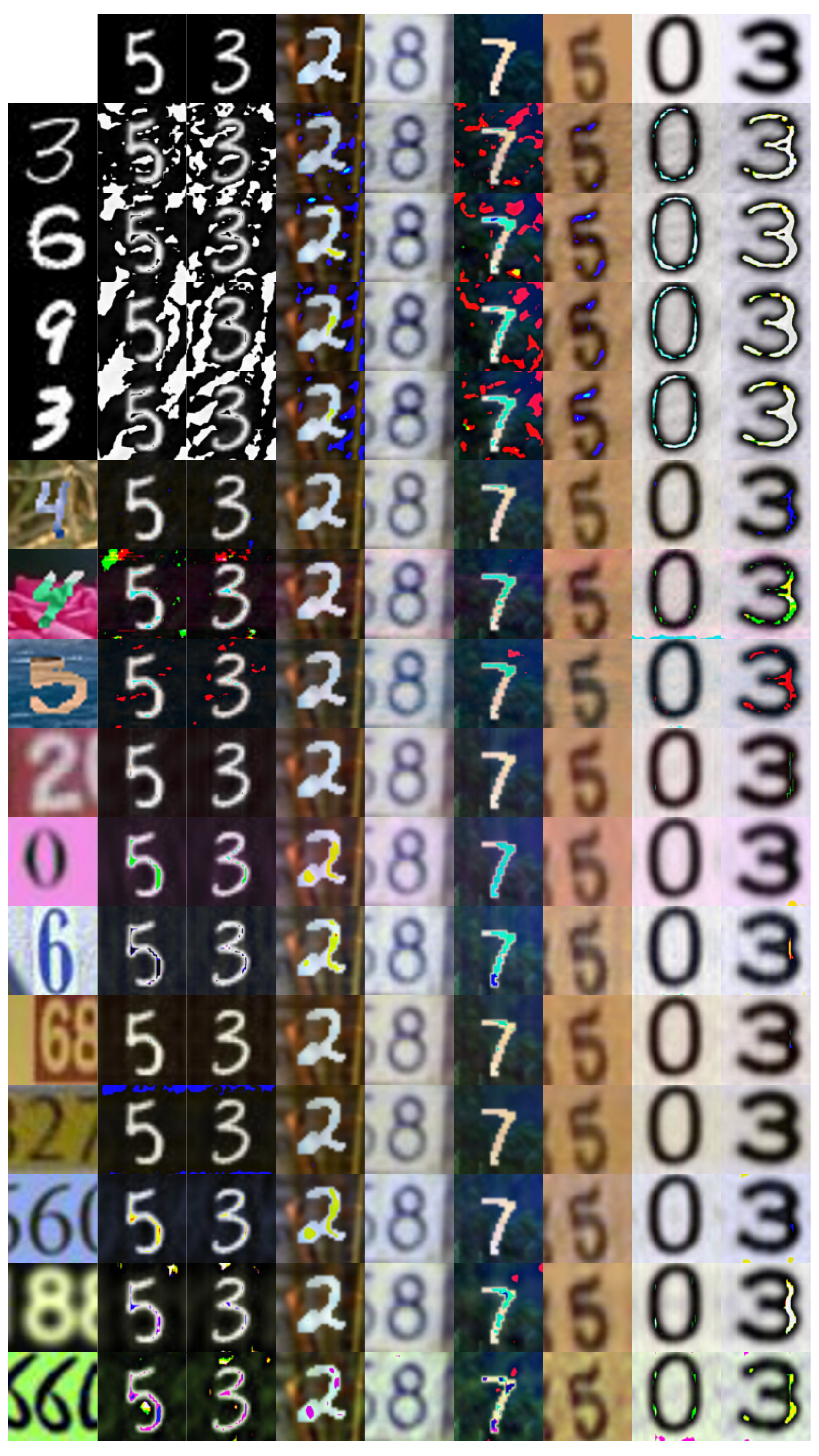}
\par\end{centering}
\caption{Stylised images produced by the Fourier-based Style Transfer of $\protect\ModelFourier$
on Digits-DG. The content and style images are visualised on the first
row and column respectively.\label{fig:Visualisations-of-diverse-Fourier-digits-dg}}
\end{figure*}

We visualise stylised images produced by the Fourier-based Style Transfer
of $\ModelFourier$ on PACS, Office-Home, and Digits-DG in Figures~\ref{fig:Visualisations-of-diverse-Fourier-PACS},~\ref{fig:Visualisations-of-diverse-Fourier-Office-Home},
and~\ref{fig:Visualisations-of-diverse-Fourier-digits-dg} respectively.

\end{document}